\pdfoutput=1
\documentclass[letterpaper]{huawei_cloud}

\usepackage{amsmath,amsfonts,bm}

\def\eqref#1{equation~\ref{#1}}

\def\1{\bm{1}}

\DeclareMathAlphabet{\mathsfit}{\encodingdefault}{\sfdefault}{m}{sl}
\SetMathAlphabet{\mathsfit}{bold}{\encodingdefault}{\sfdefault}{bx}{n}

\usepackage{amssymb}
\usepackage{enumitem}

\newcommand{\Diff}{\operatorname{Diff}}
\newcommand{\StateMatch}{\equiv_{E}}

\title{StateForge: Synthesizing Verifiable Long-Horizon Data for Tool-Using Agents via State-Change Supervision}

\author[1,*]{Wenlong~Fei}
\author[2,*]{Gangtao~Xin}
\author[3,*]{Chengyu~Shen}
\author[1]{Jiawei~Li}
\author[1]{Yedi~Hu}
\author[1]{Jingren~Zhang}
\author[2]{Hao~Jiang}
\author[2,\dagger]{Chong~Chen}
\author[1,\dagger]{Yang~Gao}

\affiliation[1]{Beijing Institute of Technology}
\affiliation[2]{Huawei Technologies Ltd.}
\affiliation[3]{Peking University}
\contribution[*]{\ Equal contribution.}
\contribution[\dagger]{\ Corresponding authors.}
\contribution{Work completed during an internship at Huawei.}

\abstract{Synthesizing long-horizon training data for tool-using agents requires reachable goals and verifiable outcomes. Successful tool calls do not guarantee task completion, while reference-call matching can reject valid alternative executions. We propose StateForge, a framework that unifies data synthesis, supervised fine-tuning (SFT), and reinforcement learning (RL) through state-change supervision. StateForge constructs executable environments with explicit state spaces and tools for querying and transforming states. Executing dependency-guided tool workflows yields reachable state-change targets, which ground task synthesis and task–target consistency checks. For each task, three trajectory synthesis modes preserve the same target: Direct provides complete goals, Progressive reveals goals in stages, and Recovery introduces recoverable environment perturbations. This shared target supports exact state-change matching to filter SFT trajectories and RL rewards that credit required changes and penalize unintended modifications. StateForge produces 544 environments and 1,697 tasks, with an average of 69 tool calls per task. With only 4B parameters, StateForge achieves state-of-the-art performance among specialized agent models on representative tool-use benchmarks, attaining 91.17\% macro-average success on $\tau^2$-Bench, 43.33\% accuracy on DeepPlanning Shopping, and a 27.25\% overall score on BFCL-v3 Multi-Turn.}

\reportlabel{}
\renewcommand{\makereportlogo}{%
  \raisebox{2.4pt}{%
    \includegraphics[trim=19bp 54bp 19bp 18bp,clip,height=9mm]{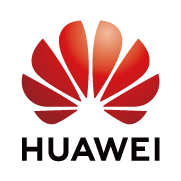}%
  }\quad
  \includegraphics[height=10.2mm]{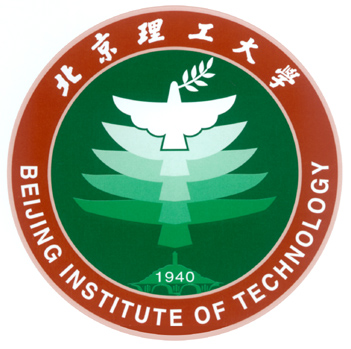}\quad
  \includegraphics[height=10.2mm]{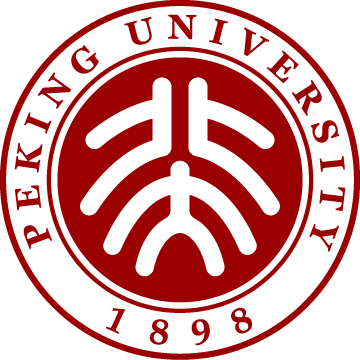}%
}

\hypersetup{
  pdftitle={\reporttitle},
  pdfauthor={Wenlong Fei, Gangtao Xin, Chengyu Shen, Jiawei Li, Yedi Hu,
    Jingren Zhang, Hao Jiang, Chong Chen, Yang Gao}
}

\begin{document}
\maketitle
\begingroup
\renewcommand{\thefootnote}{}
\footnotetext[0]{Contact us: \email{171636feiwenlong@gmail.com}}
\endgroup

\section{Introduction}
\label{sec:introduction}

Tool-using agents complete long-horizon tasks through dependent actions that transform environment states.
As these interactions grow longer, the effects of earlier actions shape the preconditions and outcomes of later ones, making local execution success an insufficient indicator of overall task completion.
Synthesizing reliable training data therefore requires reachable task objectives and trajectories whose combined effects satisfy those objectives.
Verification must account for both required changes and unintended effects, while accommodating different execution paths that realize the same target outcome.
An explicit specification of task success can thus guide task construction, trajectory selection for SFT, and outcome feedback for RL.

\begin{figure}[t]
    \centering
    \includegraphics[width=\linewidth]{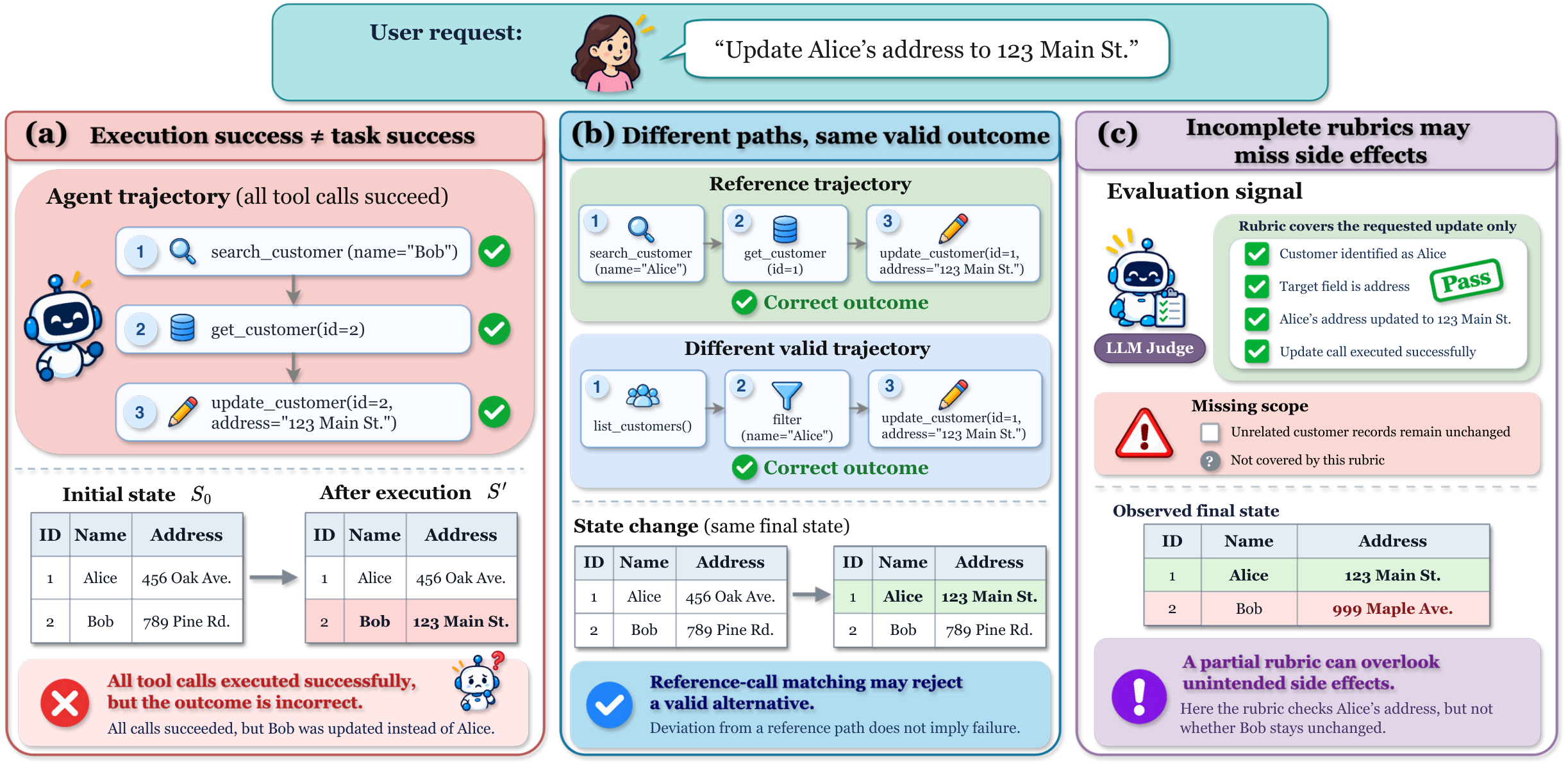}
    \caption{\textbf{Verification challenges in tool use.} For a task that requires updating Alice's address, (a) all tool calls succeed, but the agent updates Bob instead; (b) different tool-call sequences produce the same correct state change, yet reference-call matching may reject a valid alternative; and (c) a rubric covering only Alice's update can pass despite an unintended change to Bob's address.}
    \label{fig:verification-challenges}
\end{figure}

Existing methods assess synthetic data and agent behavior through execution checks, reference matching, and rubric evaluation.
Execution checks contribute to multi-stage data validation~\citep{liuzuxin2024apigen,liu2025toolace}, but execution success alone does not establish that the combined effects of tool calls fulfill the task.
Reference-call matching~\citep{xu2026covert} can penalize alternative sequences that produce the same correct state change.
Natural-language rubrics~\citep{gunjal2025rubrics,2025rubricanchors} assess task fulfillment, but omitted criteria may leave unintended modifications unchecked, while judgment errors may misclassify execution outcomes.
Figure~\ref{fig:verification-challenges} illustrates these challenges.
Environment states provide a concrete basis for assessing tool effects, and prior work uses state-based checks for verification and training feedback~\citep{song2026envscaler,xu2026envfactory}.
Building on state-based verification, we study how an execution-derived outcome specification can organize data synthesis and training.
The central challenge is to connect three requirements: task descriptions must express reachable objectives, trajectories generated under different interaction conditions must realize those objectives, and training feedback must assess the resulting effects without prescribing a reference action sequence.
Our approach uses a shared state-change target to connect these stages, grounding task generation and verification in the same executable outcome.

We propose StateForge, a framework for synthesizing verifiable long-horizon tool-use data through state-change supervision.
Within validated executable environments, we compose tool workflows based on dependencies among arguments, observations, and state preconditions.
Executing a reference workflow from an initial state \(S_0\) yields a reachable target state change \(\Delta=\operatorname{Diff}(S_0,S_g)\), where \(S_g\) is the reference execution's terminal state.
We generate a task description from the target and audit consistency: the target must cover the requested changes, and each target change must be requested or follow necessarily from the requested operations under the environment rules.
The task specifies the outcome while allowing different tool-call sequences to realize the same target.

For each task, StateForge varies the interaction conditions while preserving this target.
Direct trajectories provide the complete goal at the outset, Progressive trajectories reveal requirements in stages, and Recovery trajectories introduce recoverable environment perturbations.
These modes provide experience in planning, incorporating additional requirements, and recovering from disruptions during long-horizon interactions.
Exact state-change matching retains SFT trajectories only when all target changes are realized and no additional changes remain at termination under the environment's comparison policy.
For RL under complete goal specification, the same target defines graded outcome rewards that credit required changes and penalize unintended modifications.

StateForge produces 544 environments and 1,697 tasks, with an average of 69 tool calls per task.
We evaluate the resulting models on representative tool-use benchmarks covering constrained shopping planning, conversational service tasks, and multi-turn tool use.
With only 4B parameters, StateForge achieves state-of-the-art performance among specialized agent models, attaining 91.17\% macro-average success on $\tau^2$-Bench, 43.33\% accuracy on DeepPlanning Shopping, and a 27.25\% overall score on BFCL-v3 Multi-Turn.
Training-data comparisons and trajectory-mixture ablations further examine the downstream utility of the synthesized data and the contributions of the interaction modes.

This paper makes three contributions:
\begin{itemize}[leftmargin=*, itemsep=3pt, topsep=3pt]
    \item We develop a framework for synthesizing executable environments with rich state spaces and diverse query and update tools. Dependency-guided workflow composition enables the construction of long-horizon tasks grounded in executable tool interactions.

    \item We design Direct, Progressive, and Recovery trajectory synthesis modes to provide experience in goal-directed execution, adaptation to incrementally revealed requirements, and recovery from environment perturbations, while preserving a shared task objective.

    \item We use execution-derived state changes to unify exact-match trajectory filtering for SFT and graded outcome rewards for RL. Experiments demonstrate improvements over the corresponding backbones at both 4B and 9B scales, achieving state-of-the-art performance among specialized agent models on representative tool-use benchmarks.
\end{itemize}

\section{Related Work}
\label{sec:related}

\subsection{Tool-Use Data Synthesis}
\label{sec:related-synthesis}

Tool-use data synthesis spans executable environment construction, task generation, and multi-turn interaction synthesis.
EnvScaler~\citep{song2026envscaler} builds environments from task-derived themes, while Agent World Model~\citep{zhaoyangwang2026awm} generates code-driven, database-backed environments for agentic RL.
EnvFactory~\citep{xu2026envfactory} constructs environments from online resources, iteratively validates tool implementations and database operations, and uses dependency graphs to compose tool workflows for query generation.
APIGen~\citep{liuzuxin2024apigen} and ToolACE~\citep{liu2025toolace} generate diverse function-calling data with multiple verification steps, while APIGen-MT~\citep{prabhakar2025apigenmt} synthesizes multi-turn interactions from verified task blueprints.
Trajectory2Task~\citep{wang2026trajectory2task} converts exploratory trajectories into tasks with ambiguous, changing, or infeasible user intents.
StateForge combines execution-grounded task generation with variation in interaction conditions, using reachable state changes to guide task generation and task--target consistency auditing.
Direct, Progressive, and Recovery synthesis vary goal disclosure and environment conditions while preserving the underlying task objective.

\subsection{Training Tool-Using Agents}
\label{sec:related-training}

Training tool-using agents requires reliable demonstrations for SFT and informative rewards for RL.
Verification criteria can support both demonstration filtering and outcome feedback, drawing on reference actions, natural-language criteria, or environment states.
ToolRL~\citep{qian2025toolrl} rewards agreement with reference tool names and arguments, while COVERT~\citep{xu2026covert} preserves reference calls or answers during augmentation and uses reference matching to compute rewards for standard cases.
Rubrics as Rewards~\citep{gunjal2025rubrics} and Reinforcement Learning with Rubric Anchors~\citep{2025rubricanchors} use explicit natural-language criteria to assess task fulfillment and provide training feedback.
State-based evaluation provides a complementary basis for outcome supervision: AppWorld~\citep{trivedi2024appworld} verifies persistent effects while allowing multiple valid solutions and detecting unexpected modifications.
EnvScaler~\citep{song2026envscaler} generates terminal-state checkers from task requirements, whereas EnvFactory~\citep{xu2026envfactory} combines final-state equivalence with trajectory matching and a length penalty in its RL reward.
StateForge uses a common state-change comparison policy for demonstration selection and policy optimization.
Exact matching filters SFT trajectories, while graded RL rewards credit required effects and penalize unintended modifications.
Neither SFT filtering nor RL outcome evaluation requires agreement with a reference tool-call sequence.
\section{Problem Formulation}
\label{sec:problem}

We consider tool-use tasks and represent each task objective as a fixed target state change.
A successful trajectory must realize the required changes while preserving unrelated state, although the sequence of actions may vary.

\paragraph{Environment and interaction.}
An environment $E$ has a state space $\mathcal{S}_E$ and provides tools for reading state and modifying it within declared write scopes.
At each step, the agent selects a tool and its arguments based on the available user messages and interaction history, then receives an observation.
We denote the complete interaction trajectory by $\tau$, which includes user messages, tool calls, tool responses, and the final assistant response.

\paragraph{Task and target definition.}
We represent a task as $x=(E,q^\star,S_0,\Delta)$, where $q^\star$ is the complete task description, $S_0\in\mathcal{S}_E$ is the initial state, and $\Delta$ is the target state change.
During task construction, a reference execution from $S_0$ reaches a terminal state $S_g\in\mathcal{S}_E$, yielding a reachable target $\Delta=\operatorname{Diff}(S_0,S_g)$, where $\operatorname{Diff}$ identifies additions, deletions, and modifications between two states.
The task description expresses the objective in natural language and may be disclosed in full or in stages, while $\Delta$ remains fixed and is used for verification and reward computation without being exposed to the agent.

\paragraph{State-change verification.}
Let $S_\tau\in\mathcal{S}_E$ denote the terminal state reached under the applicable interaction conditions, including any environment interventions and subsequent agent actions.
The evaluator measures the realized state change relative to the original $S_0$ and checks whether $\operatorname{Diff}(S_0,S_\tau)\equiv_E\Delta$.
The comparison covers all state components represented in the environment schema.
The relation $\equiv_E$ denotes equivalence under a fixed comparison policy: all target changes must be realized, and no additional state changes may remain at termination.
The comparison rules, including recursive value comparison and permitted normalization of runtime-generated values, are detailed in Appendices~\ref{app:comparison-paths} and~\ref{app:normalization}.
The same comparison policy is used for task construction, trajectory validation, and reward computation.
Because verification assesses the resulting state changes rather than the tool-call sequence, different execution paths can satisfy the same target without reproducing the reference calls.
\section{Method}
\label{sec:method}

\begin{figure}[t]
    \centering
    \includegraphics[width=0.95\linewidth]{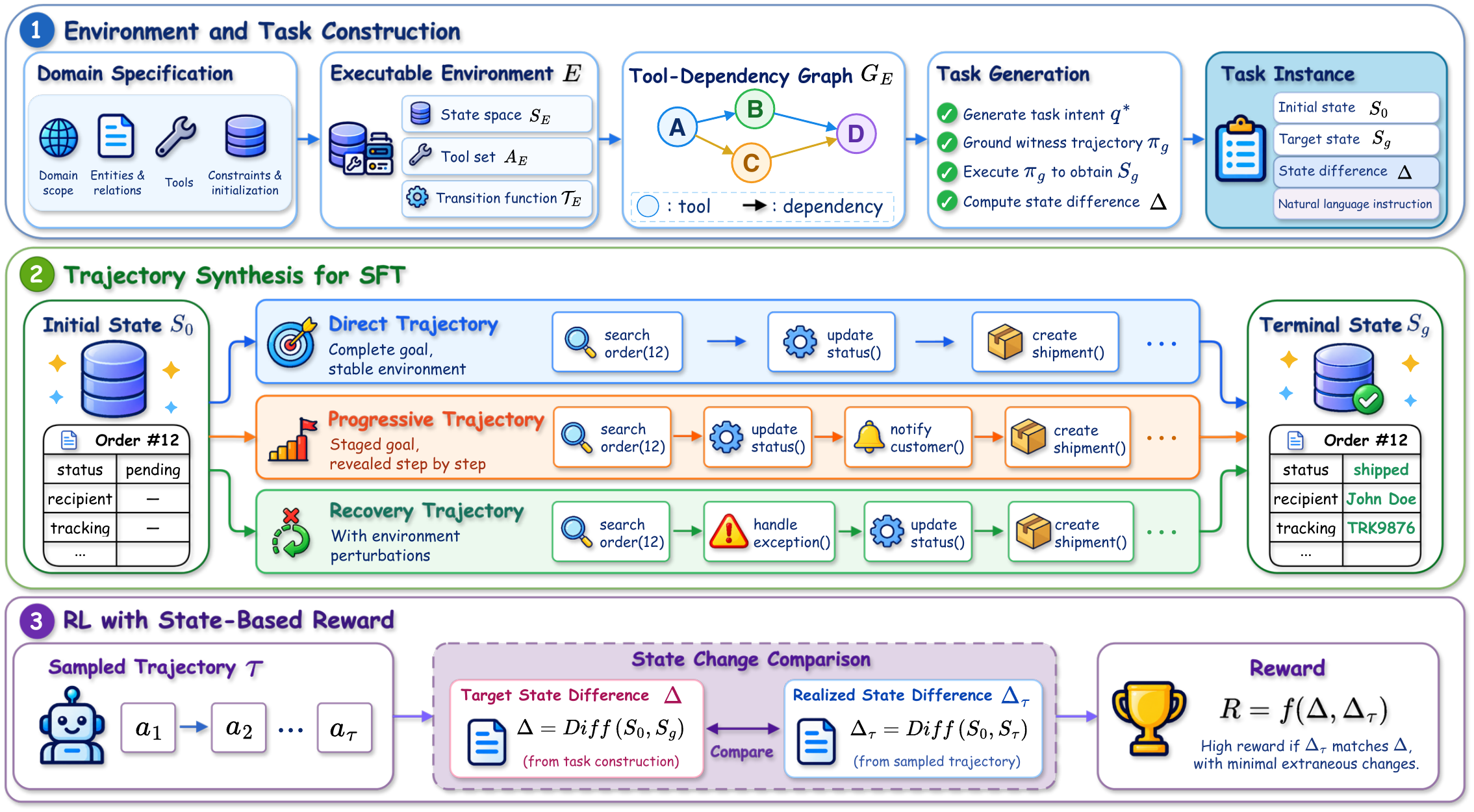}
    \caption{\textbf{Overview of StateForge.} Reference execution produces a reachable target state change $\Delta=\Diff(S_0,S_g)$, where $S_g$ is the terminal state. A task--target audit verifies consistency between the task description and $\Delta$. Direct, Progressive, and Recovery synthesis vary the interaction conditions while preserving the same target. Exact state-change matching filters SFT demonstrations, while matching provides outcome rewards for RL under the complete goal specification.}
    \label{fig:overall-pipeline}
\end{figure}

As shown in Figure~\ref{fig:overall-pipeline}, StateForge uses an execution-derived state-change target as a shared outcome specification across task construction, trajectory synthesis, and training.

\subsection{Task Construction from State Changes}
\label{sec:task-construction}

We construct long-horizon tool-use tasks by composing workflows over rich executable state spaces and deriving task objectives from state changes produced through execution.

\paragraph{Environment synthesis.}
We collect environment seeds from benchmarks, skill repositories, and other task-oriented resources and construct executable environments \(E\) with explicit state spaces \(\mathcal{S}_E\).
For each domain, we expand the state space with additional entities, fields, relations, and permitted transitions while enforcing integrity constraints.
Based on the expanded state space, we construct diverse query and update tools with explicit contracts specifying their arguments, observations, state effects, and failure conditions.
Static checks and tool-sequence execution validate initialization, state persistence, tool behavior, and complete state snapshots; environments with unresolved failures after repair are discarded.
Appendix~\ref{app:environment} details the construction and validation procedures.

\paragraph{Workflow construction.}
The tools operate on shared entities and related state transitions, creating dependencies that we represent as a directed graph \(G_E=(V_E,\mathcal{E}_E)\).
Each node represents a tool, and an edge $u\rightarrow v$ indicates that $u$ can provide an argument, identify an entity, or establish a state precondition required by $v$.
We infer candidate dependencies from tool schemas and declared behavior, then remove edges unsupported by the tool interfaces or environment constraints.
We sample workflow skeletons from \(G_E\) using bounded depth-first search to explore long dependency paths and random walks to diversify workflow structures.
We retain only candidates containing at least one state-changing operation and reduce redundancy by filtering candidates with excessive overlap in tool nodes, dependency edges, or tool-chain signatures.

\paragraph{Reference execution and target extraction.}
For each candidate workflow, we instantiate an initial state $S_0\in\mathcal{S}_E$ containing the required objects, relations, and prerequisites and validate it through the environment's initialization interface.
We then finalize the workflow as a directed acyclic graph encoding the dependencies among execution steps.
Following these dependencies, we instantiate and execute a reference tool-call sequence whose arguments are grounded in preceding observations, explicit workflow parameters, or $S_0$.
A successful reference trajectory \(\tau_r\) reaches terminal state \(S_g\), yielding a reachable candidate target \(\Delta=\operatorname{Diff}(S_0,S_g)\).

\paragraph{Task synthesis and audit.}
Even for long reference workflows, the net state change \(\Delta\) provides an explicit outcome specification from which we generate \(q^\star\), using \(S_0\) and the execution record to ground entities, context, and constraints.
The resulting task description specifies the required outcome without prescribing the reference call sequence, allowing alternative workflows.
We audit task--target consistency in both directions: every change required by $q^\star$ must be represented in $\Delta$, and every change in $\Delta$ must either be explicitly requested or necessarily follow from the requested operations under the environment rules.
The audit also checks that entity references are unambiguous and that the information required for execution is accessible to the agent.
Each accepted task is exported as $x=(E,q^\star,S_0,\Delta)$.
The initial state, target, and comparison policy remain fixed during subsequent trajectory synthesis and training.
Appendix~\ref{app:task} provides detailed procedures for workflow construction, reference execution, and task auditing.

\subsection{Trajectory Synthesis under a Shared State-Change Target}
\label{sec:trajectory-synthesis}

For each accepted task, we generate trajectories under three interaction conditions, each starting from an isolated copy of \(S_0\) and evaluated against the same target \(\Delta\).
The agent sees only disclosed task descriptions, tool schemas, and interaction history; reference terminal states, target state changes, undisclosed task descriptions, and controller plans remain hidden.

\begin{itemize}[leftmargin=*, itemsep=3pt, topsep=3pt]

    \item \textbf{Direct: complete goal specification.}
    The agent receives the complete task description \(q^\star\) at the start and interacts with the environment without intervention.
    Direct trajectories train the model to plan and execute tasks with fully specified goals.

    \item \textbf{Progressive: staged goal revelation.}
    The controller decomposes the complete task description into stage-specific descriptions $(q_1,\ldots,q_K)$ with cumulative state-change targets $\Delta^{(1)}\subsetneq\cdots\subsetneq\Delta^{(K)}=\Delta$.
    Each stage introduces additional effects from the complete target while preserving values established by earlier stages.
    Starting with $q_1$, the controller checks the current state $S$ after each action and, for $k<K$, reveals $q_{k+1}$ only when $\Diff(S_0,S)\StateMatch\Delta^{(k)}$.
    A valid decomposition requires the stage targets to be sequentially reachable from $S_0$ and to include any consequential updates required by the environment rules.
    Progressive trajectories train the model to incorporate new requirements while preserving completed effects.

    \item \textbf{Recovery: recoverable environment perturbations.}
    The agent receives the complete task description, while the controller perturbs the environment at designated trigger states, for example by overwriting a completed update or invalidating a prerequisite.
    Each intervention preserves environment invariants and keeps the original target state change $\Delta$ achievable from the perturbed state $S$.
    The controller records the intervention and exposes the resulting observation to the agent, which must adapt its execution to realize the original target.
    The resulting net state change is measured relative to $S_0$ and compared with $\Delta$, so any deviation introduced by the intervention must be repaired before termination.
    Recovery trajectories train the model to respond to environment perturbations, repair disrupted state, and replan to achieve the original objective.
\end{itemize}

Appendix~\ref{app:worked-example} presents recorded excerpts of complete and successive disclosure and a separate tool-error correction path, then details condition-specific validation.

\paragraph{Validation and SFT conversion.}
\label{sec:sft-filtering}
For each task and execution condition $\kappa$, we sample multiple candidate trajectories and record their messages, tool calls, and observations.
Let $S_\tau^\kappa=\operatorname{Exec}_{E,\kappa}(S_0,\tau)$ denote the terminal state of a candidate and $\Delta_\tau^\kappa=\Diff(S_0,S_\tau^\kappa)$ its realized state change.
We omit the superscript $\kappa$ when the interaction condition is clear.
A candidate passes validation if $
\operatorname{Accept}_x(\tau,\kappa)
=
\mathbf{1}[\Delta_\tau^\kappa\StateMatch\Delta]
\cdot
\operatorname{Valid}_x(\tau,\kappa),
$
where $\operatorname{Valid}_x(\tau,\kappa)$ captures protocol, replay, and interaction-validity checks.
Among the candidates that pass validation, we select the most efficient trajectory for each task and execution condition.
Each assistant turn in a selected trajectory forms an SFT example conditioned only on the preceding interaction history, with loss applied to that turn alone.
Appendix~\ref{app:trajectories} details the validation, selection, and conversion procedures.

\subsection{Reinforcement Learning with State-Change Rewards}
\label{sec:rl}

We use state-change matching as the primary reward signal for RL, extending the exact-match criterion used in SFT validation to provide feedback on task outcomes.
This outcome signal is computed directly from environment states, without reference-call matching or an additional reward model.

\paragraph{State-change reward.}
The state-change reward compares the state change realized by a rollout with $\Delta$ using the same comparison policy as in SFT validation.
After normalization and record alignment, let $\mathcal{P}_x$ and $\mathcal{P}_\tau$ denote the paths changed in the target and rollout under the same difference rules.
The set $\mathcal{P}_{\mathrm{match}}$ contains paths in $\mathcal{P}_x\cap\mathcal{P}_\tau$ whose resulting values or absence match the target changes.
Paths changed only by the rollout form $\mathcal{P}_{\mathrm{unexpected}}=\mathcal{P}_\tau\setminus\mathcal{P}_x$.
We assign them a penalty $U=\sum_{p\in\mathcal{P}_{\mathrm{unexpected}}}\max\{1,L_p(S_0),L_p(S_\tau)\}$, where $L_p$ counts the logical leaves under path $p$, so unexpected subtree additions or deletions are weighted by their size.
For nonempty targets with $|\mathcal{P}_x|>0$, the reward is $R_{\mathrm{goal}}=|\mathcal{P}_{\mathrm{match}}|/(|\mathcal{P}_x|+U)$.
Correctly realized target effects receive credit, while missing or incorrect effects do not, and unintended changes increase the penalty $U$.
The reward lies in \([0,1]\) and equals one if and only if the realized state change exactly matches $\Delta$ under the comparison policy.
Thus, full state-change reward requires all target effects to be realized with no additional changes remaining at termination.

\paragraph{Policy optimization.}
We optimize the policy using Group Relative Policy Optimization (GRPO; \citealp{shao2024deepseekmath}), with the state-change reward as the primary task signal.
We also use a low-weight auxiliary format reward, $R_{\mathrm{format}}\in[0,1]$, for parseable actions, structurally valid tool calls, and a valid termination format.
The final training reward is $R=0.9R_{\mathrm{goal}}+0.1R_{\mathrm{format}}$.
For each task, we sample a group of rollouts from the current policy, each starting from isolated copies of $S_0$, and compute relative advantages from the terminal rewards.
Appendix~\ref{app:reward} details the state-comparison rules, the convention for empty targets, and the auxiliary format checks.
\section{Experiments}
\label{sec:experiments}

\subsection{Experimental Setup}
\label{sec:exp-setup}

\paragraph{Training data and implementation.}
Our dataset contains 544 executable environments and 1,697 tasks. From these tasks, we construct 34,782 SFT examples from Direct, Progressive, and Recovery trajectories in a $1{:}1{:}1$ ratio.
We initialize StateForge-4B and StateForge-9B from Qwen3.5-4B and Qwen3.5-9B, respectively.
For each model scale, we first perform SFT on the constructed examples and then apply RL to the SFT checkpoint.
We denote the resulting checkpoints with the suffixes \texttt{-SFT} and \texttt{-RL}, respectively.
The SFT training parameters are provided in Appendix~\ref{app:run-settings}.

\paragraph{Benchmarks.}
\textbf{DeepPlanning Shopping}~\citep{zhang2026deepplanning} requires construction of shopping carts under product, promotion, and budget constraints. We report complete-cart accuracy (Acc.), accuracy at three constraint levels (L1--L3), and Match Score for partial product matching.
\textbf{$\tau^2$-Bench}~\citep{barres2025tau2} evaluates conversational service tasks in the Airline, Retail, and Telecom domains. We report the success rate for each domain and the unweighted mean across domains (Macro).
\textbf{BFCL-v3 Multi-Turn}~\citep{mao2024bfclv3} evaluates the Base, Missing Functions (MF), Missing Parameters (MP), and Long Context (LC) categories. We report these category scores and their unweighted mean as the multi-turn Overall score (BFCL-MT).

\paragraph{Baselines.}
We compare against proprietary models GPT-5.2 High~\citep{openai2025gpt52}, Claude Opus 4.5~\citep{anthropic2025opus45}, and Gemini-3-Pro-Pre~\citep{google2026gemini3pro}; general-purpose open-weight models DeepSeek-V3.2~\citep{deepseek2025v32} and the two Qwen3.5 backbones; and nine specialized agent models.
The agent baselines include ToolVerse-TARA-8B~\citep{zhou2026toolverse}, AWM-14B~\citep{zhaoyangwang2026awm}, EnvScaler-8B~\citep{song2026envscaler}, EnvFactory-8B~\citep{xu2026envfactory}, EnvCraft-Tool-9B~\citep{zeng2026envcraft}, ScaleEnv-32B~\citep{tu2026scaleenv}, Agent-World-14B~\citep{dong2026agentworld}, AgentSkiller-14B~\citep{sun2026agentskiller}, and TOUCAN-32B~\citep{xu2025toucan}.

\paragraph{Reporting.}
All scores are computed according to the success criteria defined by the corresponding benchmarks and are reported as percentages.
Published results on $\tau^2$-Bench are taken from the corresponding papers, whereas we evaluate the Qwen3.5 backbones and all models trained in this work.
Results on DeepPlanning Shopping and BFCL-v3 are obtained using benchmark environments configured and executed by us.
``--'' denotes results that are not reported. Score differences are computed before rounding and reported in percentage points (pp). Appendix~\ref{app:external-evaluation} gives the metric definitions.

\subsection{Main Results}
\label{sec:exp-main}

\begin{table}[!t]
\small
\caption{\textbf{Main results} (\%, $\uparrow$). Bold indicates the best result among StateForge checkpoints in each column, including ties. Shopping level scores appear in Appendix~\ref{app:shopping-difficulty}.}
\label{tab:main-results}
\footnotesize
\setlength{\tabcolsep}{2.5pt}
\resizebox{\linewidth}{!}{%
\begin{tabular}{lrrrrrrrrrrr}
\toprule
Model & \multicolumn{4}{c}{$\tau^2$-Bench} & \multicolumn{2}{c}{DeepPlanning} & \multicolumn{5}{c}{BFCL-v3 Multi-Turn} \\
\cmidrule(lr){2-5}\cmidrule(lr){6-7}\cmidrule(lr){8-12}
& Airline & Retail & Telecom & Macro & Acc. & Match & Base & MF & MP & LC & Overall \\
\midrule
\multicolumn{12}{l}{\emph{Proprietary models}} \\
GPT-5.2 High & 73.79 & 76.49 & 84.21 & 78.16 & 61.40 & 88.40 & 27.50 & 16.50 & 20.50 & 20.00 & 21.13 \\
Claude Opus 4.5 & 84.00 & 79.60 & 92.30 & 85.30 & 51.40 & 83.70 & 21.50 & 18.00 & 19.50 & 16.50 & 18.88 \\
Gemini-3-Pro-Pre & 80.50 & 75.90 & 91.00 & 82.47 & 54.00 & 83.40 & 42.00 & 40.00 & 38.50 & 32.50 & 38.25 \\
\midrule
\multicolumn{12}{l}{\emph{General-purpose open-weight models}} \\
DeepSeek-V3.2 & 63.80 & 81.10 & 96.20 & 80.37 & 54.00 & 84.00 & 35.00 & 31.50 & 24.50 & 25.00 & 29.00 \\
Qwen3.5-4B (Base) & 76.00 & 77.20 & 100.00 & 84.40 & 35.83 & 74.36 & 33.50 & 13.00 & 25.50 & 29.00 & 25.25 \\
Qwen3.5-9B (Base) & 80.00 & 86.80 & 100.00 & 88.93 & 40.83 & 77.47 & 34.50 & 8.00 & 27.50 & 30.00 & 25.00 \\
\midrule
\multicolumn{12}{l}{\emph{Agent models}} \\
ToolVerse-TARA-8B & 26.00 & 50.00 & 21.10 & 32.37 & --- & --- & --- & --- & --- & --- & --- \\
AWM-14B & 31.50 & 63.60 & 17.76 & 37.62 & 9.17 & 49.33 & 21.00 & 8.50 & 13.50 & 14.50 & 14.38 \\
EnvScaler-8B & 50.00 & 53.51 & 14.91 & 39.47 & 10.83 & 54.08 & 29.50 & 15.00 & 18.00 & 18.00 & 20.13 \\
EnvFactory-8B & 44.00 & 43.86 & 13.16 & 33.67 & 4.17 & 28.39 & 15.00 & 9.50 & 9.00 & 9.50 & 10.75 \\
EnvCraft-Tool-9B & 42.00 & 40.35 & 26.50 & 36.28 & --- & --- & --- & --- & --- & --- & --- \\
ScaleEnv-32B & 48.00 & 63.60 & 30.90 & 47.50 & --- & --- & --- & --- & --- & --- & --- \\
Agent-World-14B & 52.00 & 74.50 & 56.10 & 60.87 & --- & --- & --- & --- & --- & --- & --- \\
AgentSkiller-14B & 56.00 & 77.20 & 91.20 & 74.80 & 10.00 & 36.39 & 8.00 & 2.00 & 9.50 & 7.00 & 6.63 \\
TOUCAN-32B & 22.00 & 52.60 & 20.20 & 31.60 & 8.33 & 48.52 & 34.50 & 20.50 & 29.00 & 23.50 & 26.88 \\
\midrule
\multicolumn{12}{l}{\emph{StateForge models}} \\
StateForge-4B-SFT & 78.00 & 78.90 & 99.10 & 85.33 & 42.50 & 78.57 & 37.00 & 12.00 & \textbf{27.00} & 32.00 & 27.00 \\
StateForge-4B-RL & \textbf{84.00} & 89.50 & \textbf{100.00} & 91.17 & 43.33 & 77.84 & \textbf{39.00} & 13.50 & 26.00 & 30.50 & \textbf{27.25} \\
StateForge-9B-SFT & \textbf{84.00} & 89.47 & 98.25 & 90.57 & 45.83 & 78.21 & 36.00 & 13.50 & 23.50 & 33.50 & 26.63 \\
StateForge-9B-RL & \textbf{84.00} & \textbf{91.23} & \textbf{100.00} & \textbf{91.74} & \textbf{46.67} & \textbf{80.03} & 36.00 & \textbf{14.00} & 22.50 & \textbf{34.50} & 26.75 \\
\bottomrule
\end{tabular}}
\end{table}

\vspace{-3pt}

\paragraph{StateForge achieves strong performance with only 4B parameters.}
StateForge-4B-RL outperforms the reported specialized agent baselines on all three aggregate metrics, achieving 91.17\% $\tau^2$ Macro, 43.33\% Shopping accuracy, and 27.25\% BFCL-MT.
On BFCL-MT, it exceeds TOUCAN-32B's 26.88\% score with substantially fewer parameters.
StateForge-9B-RL further improves $\tau^2$ Macro to 91.74\%, the highest score in the comparison, and Shopping accuracy to 46.67\%.

\paragraph{Synthesized interaction data provides gains beyond model scaling.}
StateForge-4B-SFT outperforms the larger Qwen3.5-9B Base model on both Shopping accuracy and BFCL-MT.
On Shopping, training the 4B model improves accuracy by 6.67~pp, exceeding the 5.00~pp gain from increasing the backbone size from 4B to 9B.
On BFCL-MT, the two Base models perform similarly, while StateForge-4B-SFT improves over its backbone by 1.75~pp.
These results provide evidence for the benefit of training on interaction trajectories that include task execution, progressively revealed requirements, and corrective actions.

\paragraph{RL with state-change supervision provides further gains.}
RL further improves the SFT checkpoints at both model scales, increasing $\tau^2$ Macro by 5.83~pp for the 4B model and 1.17~pp for the 9B model.
These gains suggest that feedback from the policy's own executions provides useful supervision beyond the synthesized demonstrations.
By evaluating the state changes produced by the policy against the intended effects of the task, the reward links action choices to whether the required outcomes are achieved.
The gains at both model scales support the use of execution-derived state changes for continued policy optimization.

\subsection{Training Data Comparison}
\label{sec:exp-data-source}

\begin{figure}[t]
    \centering
    \includegraphics[width=\linewidth]{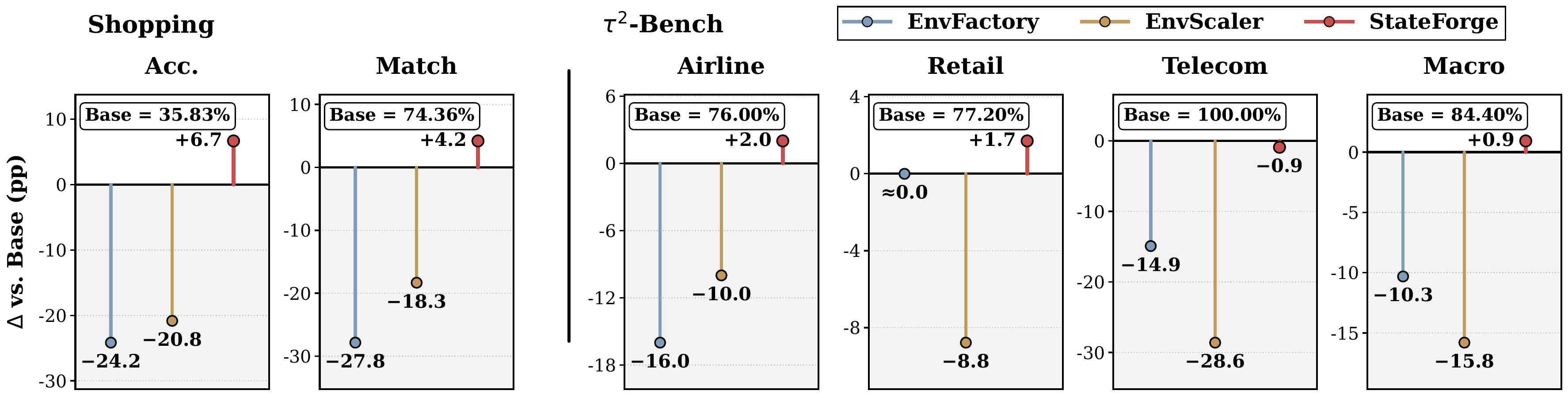}
    \caption{\textbf{Training-data comparison on Qwen3.5-4B at the SFT stage.}
    All three data sources use the same backbone and SFT configuration.
    Values show score differences from the backbone before fine-tuning
    (Base), measured in percentage points (pp).
    Positive values indicate improvements, and the zero line denotes
    unchanged performance. Insets report absolute Base scores.
    Complete absolute scores are provided in
    Appendix~\ref{app:data-source-results},
    Table~\ref{tab:data-source}.}
    \label{fig:training-data-delta}
\end{figure}

We compare the downstream utility of StateForge, EnvFactory, and EnvScaler training data using the same Qwen3.5-4B backbone at the SFT stage.
When training on EnvFactory and EnvScaler data, we use the same SFT configuration as for StateForge.
Base denotes the backbone before fine-tuning.
Table~\ref{tab:task-synthesis-comparison} summarizes the environment and tool-chain statistics of the three data sources. 
Figure~\ref{fig:training-data-delta} compares downstream score changes
relative to Base; complete absolute scores are reported in
Appendix~\ref{app:data-source-results}.

\begin{table}[t]
\centering
\caption{\textbf{Comparison of synthesized environments and tool-chain statistics.}
Task counts refer to task instances rather than SFT records. Approximate baseline lengths are marked with $\approx$.}
\label{tab:task-synthesis-comparison}
\footnotesize
\setlength{\tabcolsep}{4pt}
\renewcommand{\arraystretch}{1.15}
\begin{tabular}{@{}lcccc@{}}
\toprule
Dataset & \# Envs & \# Tasks & Avg. tools / env & Avg. tool-chain length \\
\midrule
EnvFactory~\citep{xu2026envfactory}
& 85 & 2,575 & 9.91 & $\approx 11.9$ \\
EnvScaler~\citep{song2026envscaler}
& 191 & 11,572 & 18.58 & $\approx 15$--$17$ \\
\midrule
StateForge
& 544 & 1,697 & \textbf{50.87} & \textbf{69} \\
\bottomrule
\end{tabular}
\end{table}

\vspace{-3pt}

\paragraph{StateForge data improves the utility of continued training.}
As shown in Figure~\ref{fig:training-data-delta}, StateForge is the only
training corpus that improves both Shopping accuracy and $\tau^2$ Macro
over Base under the shared backbone and SFT configuration, with gains
of 6.7~pp and 0.9~pp, respectively.
In contrast, EnvFactory reduces these scores by 24.2~pp and 10.3~pp,
while EnvScaler reduces them by 20.8~pp and 15.8~pp.
StateForge also achieves higher scores than both alternative corpora
on all six reported metrics.
Within $\tau^2$-Bench, StateForge improves Airline and Retail by
2.0~pp and 1.7~pp, respectively, with a modest 0.9~pp decrease on
Telecom; the corresponding Telecom decreases are 14.9~pp for
EnvFactory and 28.6~pp for EnvScaler.
These results show that additional interaction training does not
automatically improve a strong backbone and support the downstream
utility of the StateForge corpus under the evaluated SFT setup.

\paragraph{Broader interaction coverage and outcome verification support data quality.}
StateForge spans 544 executable environments, with an average of 50.87 tools per environment and 69 tool calls per chain.
Progressive and Recovery trajectories extend this coverage to incrementally revealed requirements and recoverable environment perturbations, while state-change verification ties retained demonstrations to the required outcomes.
The downstream gains are consistent with the utility of this combined design.
However, the comparison evaluates the complete data configuration and does not isolate individual design choices or establish data efficiency under an equal training budget.
Section~\ref{sec:exp-ablation} further examines the trajectory mixture.

\subsection{Ablation Study}
\label{sec:exp-ablation}

We ablate the StateForge trajectory mixture using the Qwen3.5-4B backbone at the SFT stage. \textbf{Direct} contains trajectories with complete task descriptions. \textbf{Direct+Progressive} adds trajectories in which task requirements are revealed across turns. \textbf{Full} additionally includes Recovery trajectories with recoverable environment perturbations. All conditions use the shared state-change target construction described in Section~\ref{sec:trajectory-synthesis}.

\begin{table}[t]
\centering
\caption{\textbf{Ablation of StateForge trajectory types on Qwen3.5-4B} (\%, $\uparrow$). Bold indicates the best result for each metric.}
\label{tab:trajectory-ablation}
\vspace{3pt}
\footnotesize
\setlength{\tabcolsep}{3pt}
\begin{tabular}{lrrrrrrrrr}
\toprule
Mixture & \multicolumn{5}{c}{Shopping} & \multicolumn{4}{c}{$\tau^2$-Bench} \\
\cmidrule(lr){2-6}\cmidrule(lr){7-10}
& L1 & L2 & L3 & Acc. & Match & Airline & Retail & Telecom & Macro \\
\midrule
Base & 40.00 & 30.00 & 40.00 & 35.83 & 74.36 & 76.00 & 77.20 & \textbf{100.00} & 84.40 \\
\midrule
Direct & 44.00 & \textbf{36.00} & 45.00 & 40.83 & \textbf{79.30} & 72.00 & 78.60 & 90.00 & 80.20 \\
Direct + Progressive & 50.00 & 34.00 & \textbf{50.00} & \textbf{43.33} & 78.75 & \textbf{82.00} & 76.30 & 96.50 & 84.93 \\
Full & \textbf{54.00} & 28.00 & \textbf{50.00} & 42.50 & 78.57 & 78.00 & \textbf{78.90} & 99.10 & \textbf{85.33} \\
\bottomrule
\end{tabular}
\end{table}

\paragraph{Progressive trajectories improve the utility of synthesized demonstrations.}
Training on Direct trajectories increases Shopping accuracy from 35.83\% to 40.83\%, but reduces the $\tau^2$ Macro score from 84.40\% to 80.20\%.
This result shows that successful execution demonstrations alone do not ensure consistent downstream gains.
Adding Progressive trajectories improves the two aggregates by 2.50~pp and 4.73~pp, respectively, bringing both above Base.
Progressive trajectories make the policy act before all requirements are available and adjust its execution as new requirements are revealed.
The policy must relate each new instruction to the current state and preserve effects that remain required.
The gains suggest that progressively revealed tasks provide useful training signals beyond complete task descriptions.

\paragraph{Recovery trajectories further improve the training mixture.}
Adding Recovery yields the highest $\tau^2$ Macro score of 85.33\%, with gains in Retail and Telecom offsetting the decrease in Airline.
These trajectories introduce recoverable deviations from the expected environment state, requiring the policy to identify unmet state-change targets and take corrective actions.
The gains are not uniform across benchmarks: Shopping accuracy decreases by 0.83~pp relative to Direct+Progressive.
Nevertheless, the full mixture exceeds Direct-only training by 1.67~pp on Shopping and 5.13~pp on $\tau^2$ Macro, and improves both aggregates over Base.
Together, these results support combining successful execution demonstrations with progressively revealed requirements and recoverable environment perturbations in StateForge.

\section{Conclusion}
\label{sec:conclusion}

We presented StateForge, a framework that uses execution-derived state changes as a unified supervision signal for task construction, trajectory synthesis, and agent training.
A shared state-change target supports the synthesis of Direct, Progressive, and Recovery trajectories, exact-match filtering for SFT, and graded state-change rewards for RL.
Experiments on representative tool-use benchmarks show improvements over the backbones at both model scales, supporting the use of state-change supervision for training tool-using agents.
Future work will extend the verification framework to accommodate multiple acceptable outcomes and constraints on intermediate behavior.
\subsection*{AI Use Statement}
Generative AI tools assisted with manuscript organization and language revision. AI-generated suggestions are not treated as scientific evidence; the authors are responsible for the technical claims, references, and final manuscript.

\subsection*{Ethics Statement}
The framework constructs tasks in synthetic environments rather than operating on live user accounts. Generated workflows can nevertheless encode biased assumptions or omit safeguards present in deployed systems. State agreement concerns modeled effects and does not establish that an action is authorized or safe in a real application. Applying the resulting agents requires appropriate permission and privacy controls.

\subsection*{Reproducibility Statement}
Sections~\ref{sec:problem} and~\ref{sec:method} define the task specification and synthesis procedure, with implementation details referenced alongside the corresponding components. 
Appendix~\ref{app:worked-example} provides recorded excerpts of Direct, Progressive, and Recovery trajectories, along with the trajectory validation, selection, and SFT conversion procedures.
Appendix~\ref{app:experiment-details} collects experimental settings and extended results.

\clearpage
\bibliographystyle{unsrtnat}
\bibliography{references}

\begin{thebibliography}{26}
\providecommand{\natexlab}[1]{#1}
\providecommand{\url}[1]{\texttt{#1}}
\expandafter\ifx\csname urlstyle\endcsname\relax
  \providecommand{\doi}[1]{doi: #1}\else
  \providecommand{\doi}{doi: \begingroup \urlstyle{rm}\Url}\fi

\bibitem[Liu et~al.(2024)Liu, Hoang, Zhang, Zhu, Lan, Kokane, Tan, Yao, Liu, Feng, Murthy, Yang, Savarese, Niebles, Wang, Heinecke, and Xiong]{liuzuxin2024apigen}
Zuxin Liu, Thai Hoang, Jianguo Zhang, Ming Zhu, Tian Lan, Shirley Kokane, Juntao Tan, Weiran Yao, Zhiwei Liu, Yihao Feng, Rithesh Murthy, Liangwei Yang, Silvio Savarese, Juan~Carlos Niebles, Huan Wang, Shelby Heinecke, and Caiming Xiong.
\newblock {APIGen}: Automated pipeline for generating verifiable and diverse function-calling datasets.
\newblock In \emph{Advances in Neural Information Processing Systems}, volume~37, 2024.
\newblock \doi{10.52202/079017-1725}.
\newblock URL \url{https://proceedings.neurips.cc/paper_files/paper/2024/hash/61cce86d180b1184949e58939c4f983d-Abstract-Datasets_and_Benchmarks_Track.html}.

\bibitem[Liu et~al.(2025)Liu, Huang, Zeng, Hao, Yu, Li, Wang, Gan, Liu, Yu, Wang, Wang, Ning, Hou, Wang, Wu, Wang, Liu, Wang, Tang, Tu, Shang, Jiang, Tang, Lian, Liu, and Chen]{liu2025toolace}
Weiwen Liu, Xu~Huang, Xingshan Zeng, Xinlong Hao, Shuai Yu, Dexun Li, Shuai Wang, Weinan Gan, Zhengying Liu, Yuanqing Yu, Zezhong Wang, Yuxian Wang, Wu~Ning, Yutai Hou, Bin Wang, Chuhan Wu, Xinzhi Wang, Yong Liu, Yasheng Wang, Duyu Tang, Dandan Tu, Lifeng Shang, Xin Jiang, Ruiming Tang, Defu Lian, Qun Liu, and Enhong Chen.
\newblock {ToolACE}: Winning the points of {LLM} function calling.
\newblock In \emph{International Conference on Learning Representations}, 2025.
\newblock URL \url{https://openreview.net/forum?id=8EB8k6DdCU}.

\bibitem[Xu et~al.(2026{\natexlab{a}})Xu, Li, Liu, Liu, Li, Shi, Zhang, Wang, Yin, Chen, Zhao, and Yin]{xu2026covert}
Siyuan Xu, Shiyang Li, Xin Liu, Tianyi Liu, Yixiao Li, Zhan Shi, Zixuan Zhang, Zilong Wang, Qingyu Yin, Jianshu Chen, Tuo Zhao, and Bing Yin.
\newblock Controllable and verifiable tool-use data synthesis for agentic reinforcement learning, 2026{\natexlab{a}}.
\newblock URL \url{https://arxiv.org/abs/2604.09813}.

\bibitem[Gunjal et~al.(2025)Gunjal, Wang, Lau, Nath, He, Liu, and Hendryx]{gunjal2025rubrics}
Anisha Gunjal, Anthony Wang, Elaine Lau, Vaskar Nath, Yunzhong He, Bing Liu, and Sean Hendryx.
\newblock Rubrics as rewards: Reinforcement learning beyond verifiable domains, 2025.
\newblock URL \url{https://arxiv.org/abs/2507.17746}.

\bibitem[Huang et~al.(2025)Huang, Zhuang, Lu, Qin, Xu, Zhao, Peng, Hu, Shen, Hu, Gu, Tu, Liu, Chen, Fu, Fan, Gu, Wang, Yang, Li, and Zhao]{2025rubricanchors}
Zenan Huang, Yihong Zhuang, Guoshan Lu, Zeyu Qin, Haokai Xu, Tianyu Zhao, Ru~Peng, Jiaqi Hu, Zhanming Shen, Xiaomeng Hu, Xijun Gu, Peiyi Tu, Jiaxin Liu, Wenyu Chen, Yuzhuo Fu, Zhiting Fan, Yanmei Gu, Yuanyuan Wang, Zhengkai Yang, Jianguo Li, and Junbo Zhao.
\newblock Reinforcement learning with rubric anchors, 2025.
\newblock URL \url{https://arxiv.org/abs/2508.12790}.

\bibitem[Song et~al.(2026)Song, Chang, Dong, Zhu, Wen, and Dou]{song2026envscaler}
Xiaoshuai Song, Haofei Chang, Guanting Dong, Yutao Zhu, Ji-Rong Wen, and Zhicheng Dou.
\newblock {EnvScaler}: Scaling tool-interactive environments for {LLM} agent via programmatic synthesis, 2026.
\newblock URL \url{https://arxiv.org/abs/2601.05808}.

\bibitem[Xu et~al.(2026{\natexlab{b}})Xu, Wang, Deng, Li, Yang, Zhu, Liu, Zhu, Huang, Chen, Deng, Mi, Shang, Zeng, and Guo]{xu2026envfactory}
Minrui Xu, Zilin Wang, Mengyi Deng, Zhiwei Li, Zhicheng Yang, Xiao Zhu, Yinhong Liu, Boyu Zhu, Baiyu Huang, Chao Chen, Heyuan Deng, Fei Mi, Lifeng Shang, Xingshan Zeng, and Zhijiang Guo.
\newblock {EnvFactory}: Scaling tool-use agents via executable environments synthesis and robust {RL}, 2026{\natexlab{b}}.
\newblock URL \url{https://arxiv.org/abs/2605.18703}.

\bibitem[Wang et~al.(2026{\natexlab{a}})Wang, Xu, Liu, Wang, Han, Yao, Yao, and He]{zhaoyangwang2026awm}
Zhaoyang Wang, Canwen Xu, Boyi Liu, Yite Wang, Siwei Han, Zhewei Yao, Huaxiu Yao, and Yuxiong He.
\newblock Agent world model: Infinity synthetic environments for agentic reinforcement learning, 2026{\natexlab{a}}.
\newblock URL \url{https://arxiv.org/abs/2602.10090}.

\bibitem[Prabhakar et~al.(2025)Prabhakar, Liu, Zhu, Zhang, Awalgaonkar, Wang, Liu, Chen, Hoang, Niebles, Heinecke, Yao, Wang, Savarese, and Xiong]{prabhakar2025apigenmt}
Akshara Prabhakar, Zuxin Liu, Ming Zhu, Jianguo Zhang, Tulika Awalgaonkar, Shiyu Wang, Zhiwei Liu, Haolin Chen, Thai Hoang, Juan~Carlos Niebles, Shelby Heinecke, Weiran Yao, Huan Wang, Silvio Savarese, and Caiming Xiong.
\newblock {APIGen-MT}: Agentic pipeline for multi-turn data generation via simulated agent-human interplay, 2025.
\newblock URL \url{https://arxiv.org/abs/2504.03601}.

\bibitem[Wang et~al.(2026{\natexlab{b}})Wang, Lu, Zhang, Chen, Dong, Huang, Gesi, Tang, Luo, Liu, Sang, Lu, Li, Lai, and Wang]{wang2026trajectory2task}
Ziyi Wang, Yuxuan Lu, Yimeng Zhang, Pei Chen, Ziwei Dong, Jing Huang, Jiri Gesi, Xianfeng Tang, Chen Luo, Qun Liu, Yisi Sang, Hanqing Lu, Manling Li, Jin Lai, and Dakuo Wang.
\newblock {Trajectory2Task}: Training robust tool-calling agents with synthesized yet verifiable data for complex user intents.
\newblock In \emph{Proceedings of the 64th Annual Meeting of the Association for Computational Linguistics (Volume 1: Long Papers)}, pages 44021--44044, 2026{\natexlab{b}}.
\newblock \doi{10.18653/v1/2026.acl-long.2037}.
\newblock URL \url{https://aclanthology.org/2026.acl-long.2037/}.

\bibitem[Qian et~al.(2025)Qian, Acikgoz, He, Wang, Chen, Hakkani-Tur, Tur, and Ji]{qian2025toolrl}
Cheng Qian, Emre~Can Acikgoz, Qi~He, Hongru Wang, Xiusi Chen, Dilek Hakkani-Tur, Gokhan Tur, and Heng Ji.
\newblock {ToolRL}: Reward is all tool learning needs.
\newblock In \emph{Advances in Neural Information Processing Systems}, volume~38, 2025.
\newblock \doi{10.52202/085713-3524}.
\newblock URL \url{https://papers.nips.cc/paper_files/paper/2025/hash/97c5b2707228e7e3fb67e4ecc2e0e607-Abstract-Conference.html}.

\bibitem[Trivedi et~al.(2024)Trivedi, Khot, Hartmann, Manku, Dong, Li, Gupta, Sabharwal, and Balasubramanian]{trivedi2024appworld}
Harsh Trivedi, Tushar Khot, Mareike Hartmann, Ruskin Manku, Vinty Dong, Edward Li, Shashank Gupta, Ashish Sabharwal, and Niranjan Balasubramanian.
\newblock {AppWorld}: A controllable world of apps and people for benchmarking interactive coding agents.
\newblock In \emph{Proceedings of the 62nd Annual Meeting of the Association for Computational Linguistics (Volume 1: Long Papers)}, pages 16022--16076, 2024.
\newblock \doi{10.18653/v1/2024.acl-long.850}.
\newblock URL \url{https://aclanthology.org/2024.acl-long.850/}.

\bibitem[Shao et~al.(2024)Shao, Wang, Zhu, Xu, Song, Bi, Zhang, Zhang, Li, Wu, and Guo]{shao2024deepseekmath}
Zhihong Shao, Peiyi Wang, Qihao Zhu, Runxin Xu, Junxiao Song, Xiao Bi, Haowei Zhang, Mingchuan Zhang, Y.~K. Li, Y.~Wu, and Daya Guo.
\newblock {DeepSeekMath}: Pushing the limits of mathematical reasoning in open language models, 2024.
\newblock URL \url{https://arxiv.org/abs/2402.03300}.

\bibitem[Zhang et~al.(2026)Zhang, Jiang, Li, Tu, Su, Deng, Guo, Lv, and Lin]{zhang2026deepplanning}
Yinger Zhang, Shutong Jiang, Renhao Li, Jianhong Tu, Yang Su, Lianghao Deng, Xudong Guo, ChenXu Lv, and Junyang Lin.
\newblock {DeepPlanning}: Benchmarking long-horizon agentic planning with verifiable constraints.
\newblock In \emph{Proceedings of the 64th Annual Meeting of the Association for Computational Linguistics (Volume 1: Long Papers)}, pages 7377--7407. Association for Computational Linguistics, 2026.
\newblock \doi{10.18653/v1/2026.acl-long.335}.
\newblock URL \url{https://aclanthology.org/2026.acl-long.335/}.

\bibitem[Barres et~al.(2025)Barres, Dong, Ray, Si, and Narasimhan]{barres2025tau2}
Victor Barres, Honghua Dong, Soham Ray, Xujie Si, and Karthik Narasimhan.
\newblock $\tau^2$-bench: Evaluating conversational agents in a dual-control environment, 2025.
\newblock URL \url{https://arxiv.org/abs/2506.07982}.

\bibitem[Mao et~al.(2024)Mao, Yan, Ji, Huang, Suresh, Huang, Yu, Gonzalez, and Patil]{mao2024bfclv3}
Huanzhi Mao, Fanjia Yan, Charlie Cheng-Jie Ji, Jason Huang, Vishnu Suresh, Yixin Huang, Xiaowen Yu, Joseph~E. Gonzalez, and Shishir~G. Patil.
\newblock {BFCL V3}: Multi-turn and multi-step function calling evaluation.
\newblock UC Berkeley Gorilla project, 2024.
\newblock URL \url{https://gorilla.cs.berkeley.edu/blogs/13_bfcl_v3_multi_turn.html}.

\bibitem[{OpenAI}(2025)]{openai2025gpt52}
{OpenAI}.
\newblock Introducing {GPT-5.2}.
\newblock Official model announcement, 2025.
\newblock URL \url{https://openai.com/index/introducing-gpt-5-2/}.

\bibitem[{Anthropic}(2025)]{anthropic2025opus45}
{Anthropic}.
\newblock Introducing {Claude Opus 4.5}.
\newblock Official model announcement, 2025.
\newblock URL \url{https://www.anthropic.com/news/claude-opus-4-5}.

\bibitem[{Google DeepMind}(2026)]{google2026gemini3pro}
{Google DeepMind}.
\newblock {Gemini 3 Pro} model card.
\newblock Official model card, 2026.
\newblock URL \url{https://deepmind.google/models/model-cards/gemini-3-pro/}.
\newblock Model released November 2025; model card updated May 2026.

\bibitem[{DeepSeek-AI}(2025)]{deepseek2025v32}
{DeepSeek-AI}.
\newblock {DeepSeek-V3.2}: Pushing the frontier of open large language models, 2025.
\newblock URL \url{https://arxiv.org/abs/2512.02556}.

\bibitem[Zhou et~al.(2026)Zhou, Yue, Hu, Shen, Zhang, Hong, Liu, and Zeng]{zhou2026toolverse}
Shuaiyu Zhou, Fengpeng Yue, Zengjie Hu, Yuanzhe Shen, Chenyang Zhang, Feng Hong, Cao Liu, and Ke~Zeng.
\newblock {ToolVerse}: Unlocking massive environments and long-horizon tasks for agentic reinforcement learning, 2026.
\newblock URL \url{https://arxiv.org/abs/2607.15660}.

\bibitem[Zeng et~al.(2026)Zeng, You, Feng, Liu, Ding, Hou, Cong, Wang, Ning, Xu, and Cai]{zeng2026envcraft}
Yirong Zeng, Shen You, Jinhang Feng, Yufei Liu, Xiao Ding, Yutai Hou, Hao Cong, Yuxian Wang, Wu~Ning, Wang Xu, and Bibo Cai.
\newblock {EnvCraft}: Synthesizing executable environments in agentic {RL} for claw-like agent, 2026.
\newblock URL \url{https://arxiv.org/abs/2609.05576}.

\bibitem[Tu et~al.(2026)Tu, Hao, Yang, Chen, Zhang, Xia, Yang, Sun, Liu, Shen, Gu, Su, and Cai]{tu2026scaleenv}
Dunwei Tu, Hongyan Hao, Hansi Yang, Yihao Chen, Yi-Kai Zhang, Zhikang Xia, Yu~Yang, Yueqing Sun, Xingchen Liu, Furao Shen, Qi~Gu, Hui Su, and Xunliang Cai.
\newblock {ScaleEnv}: Scaling environment synthesis from scratch for generalist interactive tool-use agent training, 2026.
\newblock URL \url{https://arxiv.org/abs/2602.06820}.

\bibitem[Dong et~al.(2026)Dong, Lu, Huang, Zhong, Liu, Huang, Li, Zhao, Song, Li, Jin, Zhu, Wang, Lei, Luo, Chen, Chen, Feng, Wen, and Dou]{dong2026agentworld}
Guanting Dong, Junting Lu, Junjie Huang, Wanjun Zhong, Longxiang Liu, Shijue Huang, Zhenyu Li, Yang Zhao, Xiaoshuai Song, Xiaoxi Li, Jiajie Jin, Yutao Zhu, Hanbin Wang, Fangyu Lei, Qinyu Luo, Mingyang Chen, Zehui Chen, Jiazhan Feng, Ji-Rong Wen, and Zhicheng Dou.
\newblock {Agent-World}: Scaling real-world environment synthesis for evolving general agent intelligence, 2026.
\newblock URL \url{https://arxiv.org/abs/2604.18292}.

\bibitem[Sun et~al.(2026)Sun, Ji, Cai, Wang, Wang, Li, and Chen]{sun2026agentskiller}
Zexu Sun, Bokai Ji, Hengyi Cai, Shuaiqiang Wang, Lei Wang, Guangxia Li, and Xu~Chen.
\newblock {AgentSkiller}: Scaling generalist agent intelligence through semantically integrated cross-domain data synthesis, 2026.
\newblock URL \url{https://arxiv.org/abs/2602.09372}.

\bibitem[Xu et~al.(2025)Xu, Meza~Soria, Tan, Roy, Agrawal, Poovendran, and Panda]{xu2025toucan}
Zhangchen Xu, Adriana Meza~Soria, Shawn Tan, Anurag Roy, Ashish~Sunil Agrawal, Radha Poovendran, and Rameswar Panda.
\newblock {TOUCAN}: Synthesizing 1.5m tool-agentic data from real-world {MCP} environments, 2025.
\newblock URL \url{https://arxiv.org/abs/2510.01179}.

\end{thebibliography}

\appendix
\clearpage
\section*{Appendix Overview}

The appendices follow the StateForge pipeline from environment and task construction to state-change verification, trajectory preparation, and experimental evaluation:
\begin{itemize}
    \item \textbf{Appendix~\ref{app:synthesis-details}: Environment and Task Construction Details.} Environment synthesis, tool construction, execution checks, and state coverage; workflow proposals, grounded reference execution, task-description auditing, and task export.
    \item \textbf{Appendix~\ref{app:reward}: State-Change Verification and Reward Computation.} Changed-path extraction and matching, collection comparison and runtime-value normalization, the state-change reward and its exact-matching property, the auxiliary format reward, and policy optimization.
    \item \textbf{Appendix~\ref{app:worked-example}: Trajectory Case Studies, Validation, and SFT Conversion.} Case studies of complete disclosure, successive user requests, and correction after tool errors; condition eligibility, controller events, replay and acceptance, representative selection, and information-preserving SFT conversion.
    \item \textbf{Appendix~\ref{app:experiment-details}: Experimental Details and Supplementary Results.} Training-data accounting, SFT configuration, evaluation sources and metrics, Shopping results by difficulty level, SFT and RL training-stage decomposition, and comparison controls and interpretation.
\end{itemize}

\section{Environment and Task Construction Details}
\label{app:synthesis-details}

This appendix supplements Section~\ref{sec:task-construction} with implementation and validation details for environment and task construction.

\subsection{Environment Synthesis and Validation}
\label{app:environment}

\subsubsection{Domain and State-Space Specification}
\label{app:domain-state-space}

We begin with collected task examples and retain those that require querying or updating an executable environment, excluding requests answerable from static knowledge alone. For each retained example, we infer an application setting and specify its entities, fields, relations, and operational constraints. An order-management environment, for example, may contain customers, orders, products, and inventory. Each order refers to a customer, and its items refer to products. Rules can prohibit cancellation after shipment and prevent inventory from becoming negative. We then expand the state space beyond the initial examples so that the same environment can support additional workflows and tasks.

The generator implements each application setting as a Python environment with typed state containers and an initialization interface. The schema records integrity constraints, cross-object references, and business rules. These constraints govern both initial-state construction and subsequent tool execution.

\subsubsection{Tool Construction}
\label{app:tool-construction}

Each tool specifies an exposed name, typed arguments, a return schema, a read or write designation, permitted state effects, and failure conditions. A refinement pass removes duplicate operations and decomposes overly broad ones into interfaces supporting multistep interaction. The generated methods are assembled with the environment class and checked for parseability, import consistency, and return structure.

\subsubsection{Execution Checks and Repair}
\label{app:execution-checks}

Validation first checks initialization, then executes tool sequences while recording pre-states, calls, observations, and post-states. Reads must leave state unchanged, and writes must respect their declared scopes, including consequential updates specified by their contracts. The checks also cover integrity constraints and observation structure. Invalid parameters and unmet business preconditions must produce explicit failures. The resulting state is checked against the failure contract, which need not prescribe an unchanged state.

Local repairs are followed by revalidation of the affected behavior. Environments with unresolved failures are discarded. Each retained environment packages its application description, schema, initializer, tool schemas, executable transitions, and comparison policy. These checks establish consistency on the tested interactions.

\subsection{Task Construction and Specification Audit}
\label{app:task}

\subsubsection{Workflow Proposals and Initialization}
\label{app:workflow-initialization}

For each environment, we construct the tool dependency graph $G_E$ described in Section~\ref{sec:task-construction}. An edge indicates that one tool supplies an argument, identifies an entity, or establishes a state precondition for another. For example, searching for an order can identify its ID, and reading its status can establish whether cancellation is allowed. Before sampling, we remove unknown tool identifiers, self-loops, duplicate edges, and dependencies unsupported by the interfaces or environment rules.

Sampling favors source tools and tools with many outgoing dependencies. Bounded depth-first search collects longer chains, while random walks add structural variety. Search bounds limit chain length, and higher-difficulty tasks use longer chains than lower tasks. The accepted data have an average chain length of 69 tool calls. Each candidate must include at least one state-changing operation. A main chain can be supplemented with lookup, verification, or other relevant tools. We filter candidates that overlap excessively in tool nodes, dependency edges, or chain signatures, both within and across workflow groups.

For each candidate, the state generator instantiates the required objects, relations, and prerequisites. Initialization checks reject missing containers, invalid types or enum values, and broken references. Construction controls vary dependency length, distractor density, eligibility constraints, and evidence requirements, with any state augmentation validated before reference execution.

\subsubsection{Grounded Reference Execution}
\label{app:reference-execution}

Reference calls follow the task-specific dependency order. Arguments are grounded in preceding observations, explicit workflow parameters, or the initialized state. Before execution, arguments outside the exposed schema are removed and required parameters are checked. A failed call may receive local argument repair using the tool schema, error observation, and preceding execution context. An unresolved failure invalidates the candidate.

The completed execution record and terminal state $S_g$ are retained for target extraction and task-description generation. Candidates with an empty net state change are excluded from the state-changing training set.

\subsubsection{Task Description and Bidirectional Audit}
\label{app:task-audit}

The task generator uses the target state change, initial state, and execution record to express the required outcome and ground entity descriptions and constraints. The audit follows the coverage, justification, and grounding criteria in Section~\ref{sec:task-construction}, with particular attention to two cases. First, an indirect entity description, such as a qualifying property, must identify the same object selected by the target using information available in the task description or exposed tools. Second, bookkeeping changes must be necessary consequences of the requested operations under the environment rules. Their occurrence in the reference record alone does not justify them.

Missing required effects, unsupported target effects, ambiguous references, and irreproducible targets trigger rejection or reconstruction followed by a new audit. The accepted description must specify the selected outcome sufficiently to fit the single-target formulation in Section~\ref{sec:problem}, while leaving the solution path open.

\subsubsection{Task Export and Scope}
\label{app:task-export}

Alongside the task tuple $x=(E,q^\star,S_0,\Delta)$, the export stores $S_g$, the reference record, and the comparison policy for verification and auditing. The initial state, target, and comparison policy remain fixed during subsequent trajectory synthesis and training, with agent-visible inputs restricted as specified in Section~\ref{sec:trajectory-synthesis}. State-change matching evaluates execution outcomes against this audited specification. Its reliability therefore depends on the specification's correctness.

\section{State-Change Verification and Reward Computation}
\label{app:reward}

This appendix makes the state-change verification and rollout-level reward in Sections~\ref{sec:problem} and~\ref{sec:rl} explicit. Given the initial state $S_0$, reference terminal state $S_g$, and rollout terminal state $S_\tau$, the evaluator first identifies changes relative to $S_0$, then compares the changed values, and finally computes a state-change reward and an auxiliary format reward.

\subsection{Changed Paths and Matching}
\label{app:comparison-paths}

The evaluator computes $\mathcal{P}_x=\operatorname{DiffPaths}(S_0,S_g)$ and $\mathcal{P}_\tau=\operatorname{DiffPaths}(S_0,S_\tau)$ separately. Membership in $\mathcal{P}_\tau$ establishes that a field actually changed; comparison with $S_g$ then establishes whether the new value is correct. This order matters when a rollout leaves a required field at its initial value.

\paragraph{Changed-path extraction.}
Extraction visits the union of dictionary keys and recurses over keys present on both sides. A scalar or type change contributes its current path; an added or deleted key contributes its entire subtree as one path. A changed list also contributes one path, regardless of how many elements differ. No descendant paths are added beneath an atomic subtree change. Absence is represented explicitly and is distinct from a present null value. Fields unchanged between the two states contribute no path.

\paragraph{Recursive value comparison.}
For dictionaries, $\operatorname{Equal}_E$ requires identical key sets and recursively equal values, irrespective of serialization order. Lists use the multiset comparison in Appendix~\ref{app:normalization}; ordinary scalars are compared directly under the fixed comparison policy. An added dictionary is checked in its entirety: an extra field prevents a match even though the insertion contributes only one path. Likewise, replacing a present value with null is different from deleting its key.

\paragraph{Target-path matching.}
After the generated-record alignment in Appendix~\ref{app:normalization}, both changed-path sets are expressed in the target path space. A target path matches only when the rollout changes it and its terminal value agrees with the target:
\begin{equation}
\mathcal{P}_{\mathrm{match}}
=\{p\in\mathcal{P}_x\cap\mathcal{P}_\tau:
\operatorname{Equal}_E(S_\tau[p],S_g[p])\}.
\label{eq:matched-paths}
\end{equation}
A required deletion matches only actual absence. A target path absent from $\mathcal{P}_\tau$ is a missing change; a path present in both sets but with a different resulting value is an incorrect change. Both contribute zero to the matched-path count, while the required path remains in the reward denominator. Deleting a parent instead of updating a required child produces an unexpected parent path and leaves the child unmatched.

\paragraph{Unexpected-change weighting.}
The logical leaf count $L_p(S)$ is zero for absence, one for a scalar, an empty dictionary, or a list, and the sum of child counts for a nonempty dictionary. The penalty in Section~\ref{sec:rl} is computed as
\begin{equation}
u(p)=\max\{1,L_p(S_0),L_p(S_\tau)\},
\qquad
U=\sum_{p\in\mathcal{P}_\tau\setminus\mathcal{P}_x}u(p).
\label{eq:unexpected-change-penalty}
\end{equation}
Taking the maximum over the initial and terminal subtrees accounts for both deletions and additions. An unexpected dictionary is weighted by its logical leaf count, while a list remains one comparison unit regardless of length. The lower bound of one gives every unexpected path a positive penalty. An incorrect value on a required path is already counted as an unmatched requirement; it is not also placed in $\mathcal{P}_\tau\setminus\mathcal{P}_x$.

\subsection{Value Comparison and New-Record Alignment}
\label{app:normalization}

Path membership alone does not establish a match: the evaluator must compare complete changed values and identify corresponding newly created records when their keys are generated at execution time. The following list-comparison and record-alignment rules resolve these cases before the matched-path set in Equation~\ref{eq:matched-paths} is computed.

\paragraph{Unordered lists.}
Lists are compared as multisets, retaining multiplicities and nested content. For equal-length lists $A=(a_1,\ldots,a_n)$ and $B=(b_1,\ldots,b_n)$, define
\begin{equation}
c_{ij}=\mathbf{1}[\neg\operatorname{Equal}_E(a_i,b_j)],
\qquad
C(A,B)=\min_{\pi\in\mathfrak{S}_n}\sum_{i=1}^n c_{i,\pi(i)}.
\label{eq:list-assignment}
\end{equation}
Here, $\mathfrak{S}_n$ is the set of permutations of $n$ elements. A local Hungarian solver computes the minimum-cost one-to-one assignment; lists are equal exactly when $C(A,B)=0$. Element comparisons recurse into dictionaries and nested lists without deduplication or requiring sortable elements. Unequal lengths cannot match, and two empty lists match.

Reordering alone creates no changed path. The assignment determines equality, not an element-level reward: a changed list receives credit only when the entire list matches. Order-sensitive requirements must encode order explicitly.

\paragraph{Generated record identifiers.}
When a tool creates a record, the environment may assign its key at execution time rather than taking it from the task description. The reference execution and the rollout can therefore assign different keys to records with the same task-relevant contents. A literal path comparison would then count the target addition as missing and the rollout addition as unexpected despite their equivalence. To avoid this, two dictionary records under different keys may align only if both are new relative to $S_0$ and share the same parent dictionary. For comparison, each generated key and string values within its record that exactly equal that key are replaced by \texttt{<generated-record-id>}. The complete remaining contents must agree recursively under the fixed comparison policy. Existing records and task-specified identifiers are excluded from this alignment.

Alignment is one to one: each rollout addition can satisfy at most one target addition. Matched rollout keys are mapped into the target path space before computing $\mathcal{P}_{\mathrm{match}}$ and $\mathcal{P}_{\mathrm{unexpected}}$; an unpaired target addition remains missing and an unpaired rollout addition remains unexpected. This local rule does not rename references across related records, which would require a consistent schema-level identity mapping.

\subsection{State-Change Reward and Exact Matching}
\label{app:rl-details}

The state-change reward is computed once from the terminal state of a rollout. The evaluator counts the required paths $N=|\mathcal{P}_x|$, counts the required paths both changed and correct $M=|\mathcal{P}_{\mathrm{match}}|$, and sums the weights of rollout-only paths into $U$. For a state-changing task ($N>0$), the reward in Section~\ref{sec:rl} is
\begin{equation}
R_{\mathrm{goal}}=\frac{|\mathcal{P}_{\mathrm{match}}|}{|\mathcal{P}_x|+U}
=\frac{M}{N+U}.
\label{eq:goal-reward-appendix}
\end{equation}
Each matched requirement adds one to the numerator. A missing or incorrect requirement adds nothing but remains among the $N$ required paths in the denominator; unexpected changes enlarge the denominator by $U$.

\paragraph{Exact state-change matching property.}
For $N>0$, we have $0\leq M\leq N$ and $U\geq0$, giving
\begin{equation}
0\leq\frac{M}{N+U}\leq1,
\qquad
R_{\mathrm{goal}}=1
\ \Longleftrightarrow\
(M=N\ \text{and}\ U=0)
\ \Longleftrightarrow\
\Delta_\tau\StateMatch\Delta.
\label{eq:reward-equivalence}
\end{equation}
The first equivalence follows from $M=N+U$: with $M\leq N$ and $U\geq0$, equality requires $M=N$ and $U=0$. For the second, $M=N$ means that every target path has changed to its required value, while $U=0$ excludes any unexpected path because each has positive weight. Together these conditions give the exact-matching criterion in Section~\ref{sec:problem}, under the same normalization and alignment rules.

\paragraph{Partial credit.}
For a fixed task and $U$, matching one additional target path increases the reward by $1/(N+U)$; for fixed $M>0$, increasing $U$ lowers the reward. The partial credit is per changed path: a partly correct list or a partly correct added record does not receive fractional credit within its single path.

\subsection{Auxiliary Format Reward}
\label{app:format-reward}

The auxiliary format score uses the recorded action sequence and execution diagnostics. Let $N_{\mathrm{attempt}}$ count structured actions other than the final-response action, including structured calls later marked invalid. Let $N_{\mathrm{valid}}$ count those attempts with neither a parse-error nor an invalid-action flag. Their ratio is
\begin{equation}
r_{\mathrm{call}}=
\begin{cases}
N_{\mathrm{valid}}/N_{\mathrm{attempt}},&N_{\mathrm{attempt}}>0,\\
0,&N_{\mathrm{attempt}}=0.
\end{cases}
\label{eq:format-call-ratio}
\end{equation}
A raw action that fails parsing is not a structured attempt, but it still contributes to the parse-error count. Thus a rollout with no structured tool call gets no positive call-ratio credit even if it produces a final response. The auxiliary format reward is
\begin{equation}
\begin{split}
R_{\mathrm{format}}=\operatorname{clip}_{[0,1]}\big(&r_{\mathrm{call}}
-\lambda_p n_p-\lambda_i n_i-\lambda_t I_t\\
&-\lambda_n I_n-\lambda_f I_f\big),
\end{split}
\label{eq:format-reward}
\end{equation}
Here $n_p$ and $n_i$ are the total parse-error and invalid-action counts; $I_t$, $I_n$, and $I_f$ indicate a truncated rollout, failure to enter the environment's terminated state, and failure of the final-response check, respectively. All coefficients are nonnegative. A structured invalid action reduces $r_{\mathrm{call}}$ by remaining in $N_{\mathrm{attempt}}$ but not $N_{\mathrm{valid}}$, and also contributes to $n_i$. This accounts separately for the reduced valid-call fraction and the explicit invalid-action penalty.

The final-response check requires the last structured action to be the designated communication action and its content, after trailing whitespace is removed, to end with the case-sensitive marker \texttt{Task Completed}. This marker affects $I_f$; it does not by itself establish task completion. A structurally valid tool call whose returned business result is unsuccessful can still count toward $N_{\mathrm{valid}}$, while the resulting task outcome is measured through $R_{\mathrm{goal}}$. Benchmark evaluation follows each benchmark's termination protocol.

\subsection{Policy Optimization and Reward Interpretation}
\label{app:policy-optimization}

For RL, each rollout receives the complete task description $q^\star$ and starts from an isolated copy of $S_0$. After the rollout ends, its terminal state and diagnostics yield exactly the two scores defined above. The scalar used for policy optimization is
\begin{equation}
R=0.9R_{\mathrm{goal}}+0.1R_{\mathrm{format}}.
\label{eq:combined-reward-appendix}
\end{equation}
\paragraph{Worked calculation.}
Suppose a task has two required paths. A rollout matches one, leaves the other unchanged, and adds an unrelated dictionary with two scalar leaves. Then $M=1$, $N=2$, $U=2$, and $R_{\mathrm{goal}}=1/4$. Suppose the same rollout attempts five structured non-final calls, four without error flags and one invalid, with all other penalty indicators zero. It has $r_{\mathrm{call}}=4/5$, $R_{\mathrm{format}}=\operatorname{clip}_{[0,1]}(4/5-\lambda_i)$, and $R=0.9(1/4)+0.1\operatorname{clip}_{[0,1]}(4/5-\lambda_i)$. Thus outcome credit comes from terminal-state comparison, while format credit comes from recorded action diagnostics.

For each task, the rollouts in a group are compared using these terminal scalar rewards to form relative advantages. The policy loss is applied to model-generated responses and excludes user and tool messages. Any KL regularization belongs to the optimization objective rather than the reward in Equation~\ref{eq:combined-reward-appendix}.

Equation~\ref{eq:reward-equivalence} concerns $R_{\mathrm{goal}}$. Because Equation~\ref{eq:combined-reward-appendix} also includes $R_{\mathrm{format}}$, a higher total reward need not imply exact state-change matching. SFT acceptance continues to use the exact state-change criterion in Section~\ref{sec:sft-filtering}.

\section{Trajectory Case Studies, Validation, and SFT Conversion}
\label{app:worked-example}

We illustrate Direct, Progressive, and Recovery through three trajectories in a synthetic hotel-audit task. Each panel preserves the sequence and arguments of the selected tool calls while abbreviating long messages and observations. Numbers identify steps in the trajectories, and ranges group separate calls.

\newsavebox{\trajectoryexcerptbox}
\newenvironment{trajectoryexcerpt}[1]{%
  \par\medskip\noindent
  \begin{lrbox}{\trajectoryexcerptbox}%
  \begin{minipage}{\dimexpr\linewidth-2\fboxsep-2\fboxrule\relax}%
  \small\textbf{#1}\par\smallskip\hrule\smallskip
}{%
  \end{minipage}\end{lrbox}%
  \fbox{\usebox{\trajectoryexcerptbox}}\par\medskip
}

\subsection{Direct: Grounding and Executing a Fully Disclosed Request}
\label{app:direct-trace}

The initial user message contains the full audit request, including an instruction to identify the top-ranked Amsterdam hotel, check the amenity catalog, and remove and restore a verified hotel--amenity association. The agent must first resolve the request to concrete entities before performing the prescribed update.

\begin{trajectoryexcerpt}{Direct: identifying the target hotel}
\textbf{User [0] (paraphrased).} Audit customer \texttt{USR004}'s historical hotel searches; inspect both rankings and related records; reconcile one verified amenity of the top-ranked Amsterdam hotel by removing and restoring it after checking the full catalog.

\medskip\noindent\textbf{Assistant $\rightarrow$ tool [2].} \texttt{get\_user\_search\_history}(\texttt{user\_id="USR004"}). \textbf{Tool:} search IDs \texttt{SQ001} and \texttt{SQ009}.

\noindent\textbf{Assistant $\rightarrow$ tool [4, 6].} \texttt{get\_search\_query\_by\_id}(\texttt{query\_id="SQ009"}), then \texttt{get\_city\_by\_id}(\texttt{city\_id="CTY009"}). \textbf{Tool:} \texttt{SQ009} belongs to Amsterdam.

\noindent\textbf{Assistant $\rightarrow$ tool [8, 13].} \texttt{get\_ranking\_results\_for\_query}(\texttt{query\_id="SQ009"}), then \texttt{get\_hotel\_by\_id}(\texttt{hotel\_id="HTL009"}). \textbf{Tool:} \texttt{HTL009} is rank 1; its name is Hotel Okura Amsterdam.

\medskip\noindent\emph{Omitted: other audit calls between the shown steps.}
\end{trajectoryexcerpt}

\begin{trajectoryexcerpt}{Direct: verifying and reconciling the amenity association}
\textbf{Assistant $\rightarrow$ tool [30].} \texttt{list\_all\_amenities}(). \textbf{Tool:} catalog contains \texttt{AMN001} (Wi-Fi).

\noindent\textbf{Assistant $\rightarrow$ tool [31].} \texttt{list\_amenities\_for\_hotel}(\texttt{hotel\_id="HTL009"}). \textbf{Tool:} \texttt{AMN001} is associated with the hotel.

\noindent\textbf{Assistant $\rightarrow$ tool [56].} \texttt{get\_user\_by\_id}(\texttt{user\_id="USR001"}). \textbf{Tool:} active administrator.

\medskip\noindent\emph{Omitted: other audit calls at steps 32--55 and 57--65.}

\medskip\noindent\textbf{Assistant $\rightarrow$ tool [66].} \texttt{remove\_amenity\_from\_hotel}(\texttt{hotel\_id="HTL009"}, \texttt{amenity\_id="AMN001"}, \texttt{user\_id="USR001"}). \textbf{Tool:} removal succeeds.

\noindent\textbf{Assistant $\rightarrow$ tool [67].} \texttt{add\_amenity\_to\_hotel}(\texttt{amenity\_id="AMN001"}, \texttt{current\_user\_id="USR001"}, \texttt{hotel\_id="HTL009"}). \textbf{Tool:} association succeeds.

\noindent\textbf{Assistant $\rightarrow$ tool [68].} \texttt{list\_amenities\_for\_hotel}(\texttt{hotel\_id="HTL009"}). \textbf{Tool:} \texttt{AMN001} appears again among the hotel's six amenities.
\end{trajectoryexcerpt}

\paragraph{Complete-goal planning.}
The agent grounds the initially specified audit request by resolving the customer search, city, ranking result, hotel, catalog entry, and existing association before acting. It also checks an administrator account before the write calls. The remove--restore sequence is specified from the outset, and the final read verifies the local association. This case demonstrates entity grounding and execution under complete initial disclosure.

\subsection{Progressive: A New Request After Completed Writes}
\label{app:progressive-trace}

The second trajectory begins with a request to retrieve the customer's search history. Later user turns provide operational details. Steps 84--97 show the transition from review voting to booking cleanup and the preservation of earlier effects.

\begin{trajectoryexcerpt}{Progressive: completing the review request}
\textbf{User [0] (paraphrased).} Start the audit for \texttt{USR004} by retrieving the customer's hotel search history and identifying two searches.

\medskip\noindent\emph{Intermediate audit and clarification steps omitted.}

\medskip\noindent\textbf{Assistant $\rightarrow$ user [84] (paraphrased).} Requests the acting user and review IDs for helpfulness votes; reports that flags \texttt{ML001} and \texttt{ML005} should remain.

\noindent\textbf{User [84] (paraphrased).} Vote as \texttt{USR004} on \texttt{REV002}, \texttt{REV003}, \texttt{REV024}, \texttt{REV019}, \texttt{REV020}, and \texttt{REV035}; leave both flags in place.

\noindent\textbf{Assistant $\rightarrow$ tool [85--90].} Six separate \texttt{vote\_review\_helpful} calls, one for each listed \texttt{review\_id}, all with \texttt{user\_id="USR004"}. \textbf{Tool:} all six calls report success.

\noindent\textbf{Assistant $\rightarrow$ tool [91].} \texttt{get\_review\_by\_id}(\texttt{review\_id="REV035"}). \textbf{Tool:} \texttt{helpfulness\_votes=10}; an earlier read had returned 9.
\end{trajectoryexcerpt}

\begin{trajectoryexcerpt}{Progressive: the subsequent booking request}
\textbf{Assistant $\rightarrow$ user [92] (paraphrased).} Reports the six votes and preservation of \texttt{ML001} and \texttt{ML005}.

\noindent\textbf{User [92] (paraphrased).} Now cancel the discovered pending bookings that belong to their booking owners and are suitable for cleanup.

\medskip\noindent\textbf{Assistant $\rightarrow$ tool [93--94].} \texttt{get\_booking\_by\_id}(\texttt{booking\_id="BKG002"}) and \texttt{get\_booking\_by\_id}(\texttt{booking\_id="BKG018"}). \textbf{Tool:} both are \texttt{pending}, owned by \texttt{USR005}, with empty \texttt{payment\_id}.

\noindent\textbf{Assistant $\rightarrow$ tool [95--96].} Two \texttt{cancel\_booking} calls, for \texttt{BKG002} and \texttt{BKG018}, each with \texttt{user\_id="USR005"}. \textbf{Tool:} both succeed.

\noindent\textbf{Assistant $\rightarrow$ tool [97].} \texttt{get\_booking\_by\_id}(\texttt{booking\_id="BKG002"}). \textbf{Tool:} \texttt{status="cancelled"}.
\end{trajectoryexcerpt}

\begin{trajectoryexcerpt}{Progressive: cumulative state changes}
\textbf{Review vote counts, prior reads $\rightarrow$ terminal state:} \texttt{REV002} $8\to9$; \texttt{REV003} $15\to16$; \texttt{REV024} $19\to20$; \texttt{REV019} $8\to9$; \texttt{REV020} $13\to14$; \texttt{REV035} $9\to10$. The prior values appear in steps 12, 13, 69, and 70.

\noindent\textbf{Booking status, prior reads $\rightarrow$ terminal state:} \texttt{BKG002} and \texttt{BKG018} both \texttt{pending}$\to$\texttt{cancelled}. Moderation flags \texttt{ML001} and \texttt{ML005} remain in place.

\end{trajectoryexcerpt}

\paragraph{Cumulative effects across user requests.}
The booking request follows the assistant's review-completion message at step 92. The agent checks both bookings before canceling them under their recorded owner, \texttt{USR005}, while the terminal state retains the earlier vote increments. The case thus exhibits successive disclosure and accumulation of the listed effects.

\subsection{Recovery: Repairing a Disrupted Amenity Operation}
\label{app:recovery-trace}

To construct the Recovery trajectory, we manually introduce a disruption at the \texttt{HTL009}--\texttt{AMN001} amenity-removal step. The first two removal attempts receive permission-denied responses, interrupting the planned remove--restore sequence. The agent checks account roles, changes the acting user, and completes the original request.

\begin{trajectoryexcerpt}{Recovery: controller-induced authorization disruption}
\textbf{Assistant $\rightarrow$ tool [38].} \texttt{list\_amenities\_for\_hotel}(\texttt{hotel\_id="HTL009"}). \textbf{Tool:} \texttt{AMN001} is present.

\noindent\textbf{Assistant $\rightarrow$ tool [39].} \texttt{list\_all\_amenities}(). \textbf{Tool:} catalog contains \texttt{AMN001} (Wi-Fi).

\medskip\noindent\textbf{Controller [40].} Before the next agent action, the controller perturbs the authorization state by changing \texttt{USR004}'s role from \texttt{admin} to \texttt{customer}. The intervention is not directly disclosed to the agent.

\noindent\textbf{Assistant $\rightarrow$ tool [41].} \texttt{remove\_amenity\_from\_hotel}(\texttt{hotel\_id="HTL009"}, \texttt{amenity\_id="AMN001"}, \texttt{user\_id="USR004"}). \textbf{Tool:} failure; \texttt{USR004} now has role \texttt{customer} and lacks permission.

\noindent\textbf{Assistant $\rightarrow$ tool [42].} \texttt{get\_user\_by\_id}(\texttt{user\_id="USR002"}). \textbf{Tool:} active moderator.

\noindent\textbf{Assistant $\rightarrow$ tool [43].} \texttt{remove\_amenity\_from\_hotel}(\texttt{hotel\_id="HTL009"}, \texttt{amenity\_id="AMN001"}, \texttt{user\_id="USR002"}). \textbf{Tool:} failure; the moderator also lacks permission.

\noindent\textbf{Assistant $\rightarrow$ tool [44].} \texttt{get\_user\_by\_id}(\texttt{user\_id="USR001"}). \textbf{Tool:} \texttt{role="admin"}, \texttt{status="active"}.
\end{trajectoryexcerpt}

\begin{trajectoryexcerpt}{Recovery: revised action and state restoration}
\textbf{Assistant $\rightarrow$ tool [45].} \texttt{remove\_amenity\_from\_hotel}(\texttt{hotel\_id="HTL009"}, \texttt{amenity\_id="AMN001"}, \texttt{user\_id="USR001"}). \textbf{Tool:} removal succeeds.

\noindent\textbf{Assistant $\rightarrow$ tool [46].} \texttt{add\_amenity\_to\_hotel}(\texttt{hotel\_id="HTL009"}, \texttt{amenity\_id="AMN001"}, \texttt{current\_user\_id="USR001"}). \textbf{Tool:} association succeeds.

\medskip\noindent\textbf{Controller [47].} After the recovery sequence, the controller restores the perturbed authorization state by changing \texttt{USR004}'s role from \texttt{customer} back to \texttt{admin}.

\medskip\noindent\emph{Omitted: other audit calls at steps 48--62.}

\medskip\noindent\textbf{Assistant $\rightarrow$ tool [63].} \texttt{list\_amenities\_for\_hotel}(\texttt{hotel\_id="HTL009"}). \textbf{Tool:} \texttt{AMN001} is again present.

\noindent\textbf{Assistant $\rightarrow$ tool [64].} \texttt{get\_user\_by\_id}(\texttt{user\_id="USR004"}). \textbf{Tool:} \texttt{role="admin"}, confirming that the controller-induced authorization change has been reverted.
\end{trajectoryexcerpt}

\paragraph{Recovery after disruption.}
The controller-induced intervention temporarily changes the authorization state of the user initially selected for the removal, causing the next tool call to fail. The agent observes the permission failure, checks alternative user roles, unsuccessfully tries an active moderator, and then identifies the active administrator \texttt{USR001}. Reissuing the removal with valid credentials succeeds, after which the requested amenity association is restored. The controller then reverts its temporary authorization perturbation, and the final reads verify both the requested amenity state and restoration of the unrelated user state. The trajectory therefore realizes the disruption--observation--revision--recovery--verification pattern while leaving no residual controller-induced state change.

\subsection{Trajectory Construction and SFT Conversion}
\label{app:trajectories}

For the synthesis procedure in Section~\ref{sec:trajectory-synthesis}, the runner records user messages, assistant actions, tool observations, state snapshots, termination status, and condition-specific controller events for replay.

\subsubsection{Condition Eligibility and Controller Events}
\label{app:condition-eligibility}

\paragraph{Direct.}
The complete task description is placed in the initial user message, and the designated communication action ends the interaction.

\paragraph{Progressive.}
Stage targets use strict inclusion of required path--value effects under the fixed comparison policy. Each stage must be reachable from the preceding cumulative state through the exposed tools, including any consequential updates. For example, if the first stage fixes a counter to one and the second necessarily increments it to two, the stages cannot form the required nested targets. The same restriction applies to an atomic list whose required value would change between stages. Such effects must be grouped into a compatible stage rather than omitted from comparison.

Each reveal record stores the stage index, state snapshot, and gate result. After the current gate passes, the newly revealed description is appended to the interaction history; future descriptions and cumulative targets remain hidden. Clarifications and read-only exchanges may occur within a stage without defining an additional state-change target.

\paragraph{Recovery.}
An intervention specifies a trigger, a state transformation, and the resulting agent-visible observation, and is applied at a declared boundary between agent actions. It disrupts a completed update or a prerequisite while satisfying the invariant and reachability requirements in Section~\ref{sec:trajectory-synthesis}.

At the actual trigger state, a recovery witness is executed on an isolated perturbed-state copy to verify that the full target $\Delta$ remains achievable relative to $S_0$. The witness is retained for auditing but is not exposed to the agent or executed on the live trajectory. The event record stores the trigger outcome, pre- and post-intervention states, agent-visible observation, and reachability-check result. A rollout is retained as a Recovery candidate only if an intervention was applied.

This check includes consequential effects of the repair. For example, restoring a deleted association is insufficient if the repair necessarily leaves an additional audit record outside $\Delta$. In that case, the intervention does not satisfy the reachability requirement even though the association itself can be restored.

\subsubsection{Replay and Acceptance}
\label{app:replay-acceptance}

Message processing preserves call--response associations and rejects unsupported action structures. Changes to the executable history require replay from $S_0$, including recorded disclosures and interventions, with their preconditions and observations checked for consistency.

Record and protocol checks establish whether a trajectory is a valid execution under the specified interaction condition. Progressive records must complete disclosure gates in order, and Recovery records must contain an applied admissible intervention. Applicable response obligations and interaction constraints are checked against the recorded history. Records with failed or undefined replay are excluded.

For eligible candidates, the verifier computes $\Diff(S_0,S_\tau^\kappa)$ from the replayed terminal state and applies $\operatorname{Accept}_x(\tau,\kappa)$ from Section~\ref{sec:trajectory-synthesis}. This keeps record and protocol checks separate from the outcome criterion. A temporary execution error need not cause rejection if its state effects are repaired and the interaction requirements remain satisfied.

\subsubsection{Representative Selection}
\label{app:representative-selection}

For each task and condition, we retain the most efficient accepted trajectory, as specified in Section~\ref{sec:sft-filtering}. A singleton pool contributes its sole candidate; an empty pool contributes none. Selection therefore produces at most one retained trajectory per task and condition before conversion into turn-level training records.

\subsubsection{Information-Preserving SFT Conversion}
\label{app:sft-conversion}

The converter applies the target model's chat template to structured messages and tools. For a retained assistant turn $a_t$, the training example pairs its preceding interaction history $H_t$ with $a_t$ as the supervised output. Loss is applied only to $a_t$; system and user messages, tool observations, and earlier assistant turns provide context without contributing loss in that example.

For successive assistant turns $a_t$ and $a_{t+1}$ separated by a tool observation $o_t$, the next example uses $H_{t+1}=(H_t,a_t,o_t)$ and supervises only $a_{t+1}$. Newly disclosed requirements or perturbation observations enter the history at their recorded positions. Controller-only targets, validation witnesses, and undisclosed requirements are excluded. Thus, one retained trajectory yields multiple SFT records, each preserving the information available at its supervised turn.

If tool or parameter identifiers are anonymized, the same mapping is applied to schemas, calls, and associated messages while preserving argument meaning and call--response identity. Original execution records remain available for replay.

\section{Experimental Details and Supplementary Results}
\label{app:experiment-details}

This appendix provides the SFT configuration, clarifies how training
records and evaluation aggregates are counted, and reports complete
training-data comparison scores, Shopping results by difficulty
level, and training-stage differences.

\subsection{Training-Record Granularity}
\label{app:training-data}

StateForge counts each supervised assistant turn as one SFT record, following the conversion in Appendix~\ref{app:sft-conversion}. The recorded EnvScaler and EnvFactory inventories contain approximately 9K and 26.5K SFT records, respectively~\citep{song2026envscaler,xu2026envfactory}. These counts do not establish equal token exposure: record granularity and interaction length may differ across sources, and the shared epoch count does not match supervised tokens or optimizer updates.

\subsection{SFT Configuration}
\label{app:run-settings}

Table~\ref{tab:sft-settings} gives the Qwen3.5-4B SFT configuration used for all three data sources in Section~\ref{sec:exp-data-source}. StateForge-9B shares the listed optimizer, learning rate, schedule, warmup ratio, epoch count, sequence length, GPU count, and gradient accumulation.

\begin{table}[htbp]
\centering
\caption{Qwen3.5-4B SFT configuration, shared across all three data sources in the training-data comparison.}
\label{tab:sft-settings}
\small
\begin{tabular}{@{}p{0.30\linewidth}p{0.62\linewidth}@{}}
\toprule
Setting & Value \\
\midrule
Training framework & LLaMA-Factory \\
Fine-tuning method & Full-parameter language fine-tuning; vision tower and multimodal projector frozen \\
Optimizer & PyTorch AdamW \\
Learning rate & $1\times10^{-6}$ \\
Training epochs & 3 \\
Maximum sequence length & 32k tokens \\
Number of GPUs & 8 \\
Per-device batch size & 1 \\
Gradient accumulation & 32 steps \\
Effective batch size & 256 training examples per optimizer update \\
Learning-rate schedule & Cosine decay \\
Warmup ratio & 0.1 \\
Maximum gradient norm & 1.0 \\
Chat template & \texttt{qwen3\_5}, thinking enabled \\
History masking & Enabled \\
Sequence packing & Disabled \\
Numerical precision & bfloat16 \\
Memory and attention & Gradient checkpointing, FlashAttention-2, DeepSpeed \\
\bottomrule
\end{tabular}
\end{table}

\subsection{Evaluation Aggregation}
\label{app:external-evaluation}

Shopping Acc. measures complete-cart correctness, while Match Score credits partial product matching under the DeepPlanning evaluator~\citep{zhang2026deepplanning}. We use the evaluator's aggregate Acc. directly; the displayed L1--L3 scores are separate breakdowns and are not used to reconstruct it. The BFCL-MT Overall column averages only the four multi-turn categories named in Section~\ref{sec:exp-setup}, so it should not be read as the full BFCL-v3 leaderboard score.

\subsection{Complete Training-Data Comparison Results}
\label{app:data-source-results}

\begin{table}[t]
\centering
\caption{\textbf{Complete training-data comparison on Qwen3.5-4B
at the SFT stage} (\%, $\uparrow$). 
Absolute scores corresponding to Figure~\ref{fig:training-data-delta}. 
Base denotes the backbone before fine-tuning. 
Bold indicates the best fine-tuned result in each column.}
\label{tab:data-source}
\small
\setlength{\tabcolsep}{3.8pt}
\begin{tabular}{lrrrrrr}
\toprule
Training data & \multicolumn{2}{c}{Shopping} & \multicolumn{4}{c}{$\tau^2$-Bench} \\
\cmidrule(lr){2-3}\cmidrule(lr){4-7}
& Acc. & Match & Airline & Retail & Telecom & Macro \\
\midrule
Base & 35.83 & 74.36 & 76.00 & 77.20 & 100.00 & 84.40 \\
EnvFactory data & 11.67 & 46.52 & 60.00 & 77.19 & 85.09 & 74.09 \\
EnvScaler data & 15.00 & 56.04 & 66.00 & 68.42 & 71.39 & 68.60 \\
StateForge data & \textbf{42.50} & \textbf{78.57} & \textbf{78.00} & \textbf{78.90} & \textbf{99.10} & \textbf{85.33} \\
\bottomrule
\end{tabular}
\end{table}

Table~\ref{tab:data-source} reports the complete absolute scores
underlying Figure~\ref{fig:training-data-delta}.
All three training corpora are evaluated using the same Qwen3.5-4B
backbone and SFT configuration. Base denotes the backbone before
fine-tuning.

\subsection{Shopping Results by Difficulty Level}
\label{app:shopping-difficulty}

Table~\ref{tab:main-breakdown} breaks down the Shopping evaluations in Table~\ref{tab:main-results} by constraint level, omitting models without reported level scores.

\begin{table}[htbp]
\centering
\caption{Shopping accuracy by difficulty level (\%). Aggregate Acc. and Match scores appear in Table~\ref{tab:main-results}. A dash in Stage indicates an external baseline.}
\label{tab:main-breakdown}
\small
\begin{tabular}{@{}llrrr@{}}
\toprule
Model & Stage & L1 & L2 & L3 \\
\midrule
AWM-14B & --- & 16.00 & 6.00 & 0.00 \\
EnvScaler-8B & --- & 16.00 & 10.00 & 0.00 \\
EnvFactory-8B & --- & 8.00 & 2.00 & 0.00 \\
AgentSkiller-14B & --- & 14.00 & 8.00 & 5.00 \\
TOUCAN-32B & --- & 16.00 & 4.00 & 0.00 \\
\midrule
Qwen3.5-4B & Base & 40.00 & 30.00 & 40.00 \\
StateForge-4B-SFT & SFT & 54.00 & 28.00 & 50.00 \\
StateForge-4B-RL & RL & 54.00 & 32.00 & 45.00 \\
\midrule
Qwen3.5-9B & Base & 46.00 & 36.00 & 40.00 \\
StateForge-9B-SFT & SFT & 52.00 & 36.00 & 55.00 \\
StateForge-9B-RL & RL & 56.00 & 38.00 & 45.00 \\
\bottomrule
\end{tabular}
\end{table}

At both sizes, SFT improves L1 and L3 over Base, but L2 falls at 4B and remains unchanged at 9B. Continued RL improves L2 and reduces L3 at both sizes; L1 remains unchanged at 4B and improves at 9B. Thus, the aggregate gains in Table~\ref{tab:main-results} do not imply gains at every constraint level.

Table~\ref{tab:data-source-levels} further breaks down the Shopping
results from Table~\ref{tab:data-source} by difficulty level.

\begin{table}[htbp]
\centering
\caption{Shopping level accuracy for the training-data comparison on Qwen3.5-4B at the SFT stage (\%). Base denotes the backbone before fine-tuning.}
\label{tab:data-source-levels}
\small
\begin{tabular}{@{}lrrr@{}}
\toprule
Training data & L1 & L2 & L3 \\
\midrule
Base & 40.00 & 30.00 & 40.00 \\
EnvFactory & 20.00 & 8.00 & 0.00 \\
EnvScaler & 18.00 & 14.00 & 10.00 \\
StateForge & 54.00 & 28.00 & 50.00 \\
\bottomrule
\end{tabular}
\end{table}

StateForge exceeds both alternative training corpora at each level. Relative to Base, it improves L1 and L3, while L2 decreases by 2~pp.

\subsection{Training-Stage Decomposition}
\label{app:additional-rl}

Table~\ref{tab:stage-deltas} separates SFT gains, additional RL gains, and total gains over Base for the checkpoints in Table~\ref{tab:main-results}.

\begin{table}[htbp]
\centering
\caption{Training-stage differences (pp), computed before rounding. Positive values indicate improvement. SFT and RL denote the corresponding StateForge checkpoint suffixes; Base denotes the Qwen3.5 backbone of the same size.}
\label{tab:stage-deltas}
\small\setlength{\tabcolsep}{4pt}
\begin{tabular}{@{}llrrrr@{}}
\toprule
Size & Comparison & Shopping Acc. & Match & $\tau^2$ Macro & BFCL-MT \\
\midrule
4B & SFT minus Base & 6.67 & 4.21 & 0.93 & 1.75 \\
4B & RL minus SFT & 0.83 & -0.73 & 5.83 & 0.25 \\
4B & RL minus Base & 7.50 & 3.48 & 6.77 & 2.00 \\
\midrule
9B & SFT minus Base & 5.00 & 0.74 & 1.64 & 1.63 \\
9B & RL minus SFT & 0.84 & 1.82 & 1.17 & 0.13 \\
9B & RL minus Base & 5.84 & 2.56 & 2.81 & 1.75 \\
\bottomrule
\end{tabular}
\end{table}

At 4B, RL increases Shopping Acc. by 0.83~pp while reducing Match Score by 0.73~pp. The net $\tau^2$ Macro gains over Base come from Airline and Retail, since Telecom is at 100\% for both Base and RL checkpoints at each size.

\makeatletter
\setlength{\@fptop}{0pt}
\makeatother
\end{document}